\documentclass[letterpaper, 10 pt, conference]{ieeeconf}  % Comment this line out if you need a4paper

\IEEEoverridecommandlockouts                              % This command is only needed if 
\usepackage{color}
\usepackage{amsmath} % assumes amsmath package installed
\usepackage{amssymb}  % assumes amsmath package installed
\usepackage[table,xcdraw]{xcolor}
\usepackage{graphicx}
\usepackage{booktabs} % For better horizontal lines
\usepackage{tabularx} % For controlling table width
\usepackage{array}    % For better column alignment
\usepackage{multirow}
\usepackage{stfloats}
\usepackage{caption}
\usepackage{subcaption}
\usepackage{cite}
\usepackage{algorithm}
\usepackage{algpseudocode}
\usepackage{comment}

\usepackage[whole]{bxcjkjatype}

\title{\LARGE \bf
Experience-Based Feasibility-Aware Generative Adversarial Imitation from Observation under Embodiment Mismatch
}

\author{
Yoshiki Takebayashi$^{1}$, Giovanni Perantoni$^{1}$, Hikaru Sasaki$^{1}$, Matteo Saveriano$^{2}$, Takamitsu Matsubara$^{1}$
\thanks{*This work was supported by JSPS KAKENHI Grant Numbers 26K17349 and 24K03018.
} % <-this % stops a space
\thanks{$^{1}$Y. Takebayashi, G. Perantoni, H. Sasaki, and T. Matsubara are with the Division of Information Science, Graduate School of Information Science, Nara Institute of Science and Technology (NAIST), Nara, Japan.
email: {\tt\small takebayashi.yoshiki.ub0@naist.ac.jp, perantoni.giovanni.pm7@naist.ac.jp, sasaki.hikaru@is.naist.jp, takam-m@is.naist.jp}}
\thanks{$^{2}$M. Saveriano is with the Department of Industrial Engineering of the University of Trento, Trento, Italy.
email: {\tt\small matteo.saveriano@unitn.it}}
}

\begin{document}

\maketitle
\thispagestyle{empty}
\pagestyle{empty}

%%%%%%%%%%%%%%%%%%%%%%%%%%%%%%%%%%%%%%%%%%%%%%%%%%%%%%%%%%%%%%%%%%%%%%%%%%%%%%%%
\begin{abstract}
With the increasing use of robot-free demonstration interfaces that provide state trajectories without action labels, imitation from observation has become a promising approach for learning robot behaviors from human demonstrations. However, due to differences in embodiment and dynamics between humans and robots, demonstrated human motions may not be feasible for the robot, potentially degrading policy performance. In this study, we propose Experience-Based Feasibility-Aware Generative Adversarial Imitation from Observation (EF-GAIfO), which estimates the feasibility of state-only demonstrations from the robot's own experience rather than relying on explicit dynamics models or large prior exploration datasets. A key feature of EF-GAIfO is that the notion of feasibility evolves with policy learning: as the policy improves and the robot experiences a broader range of state transitions, the feasible region is progressively expanded, allowing additional demonstrations to be incorporated into learning.
This enables feasibility-aware imitation that adapts to the current stage of policy learning, rather than relying on a pre-designed feasibility criterion.
We validate the effectiveness of EF-GAIfO on a locomotion task in simulation and on a real quadruped robot performing a object-reaching-and-grasping task.
\end{abstract}

%%%%%%%%%%%%%%%%%%%%%%%%%%%%%%%%%%%%%%%%%%%%%%%%%%%%%%%%%%%%%%%%%%%%%%%%%%%%%%%
%%%%%%%%%%%%%%%%%%%%%%%%%%%%%%%%%%%%%%%%%%%%%%%%%%%%%%%%%%%%%%%%%%%%%%%%%%%%%%%
%%%%%%%%%%%%%%%%%%%%%%%%%%%%%%%%%%%%%%%%%%%%%%%%%%%%%%%%%%%%%%%%%%%%%%%%%%%%%%%
\section{Introduction}
In recent years, imitation learning approaches using human demonstrations collected through robot-free demonstration interfaces have attracted increasing attention. 
Several approaches have been proposed, ranging from purely video-based ones \cite{young2021visual} to those based on external infrastructures \cite{ohkawa2026yubi, sasaki2026robotic}, and those based on self-contained onboard tracking \cite{ha2025UMI, shafiullah2023bringing}.
Such interfaces allow users to provide demonstrations through their own hand movements without directly teleoperating a robot.
However, demonstrations collected through such interfaces typically consist of state trajectories without corresponding robot action labels.
Imitation from Observation (IfO) \cite{torabi2018bco, torabi2019}, which learns policies from state-only demonstrations, is therefore a suitable framework for utilizing such data.

However, due to an embodiment mismatch between humans and robots, demonstrated state transitions collected via robot-free demonstration interfaces are not necessarily feasible for the robot.
Such demonstrations may contain state transitions that cannot be reproduced under the robot's embodiment.
If these demonstrations are directly used for IfO, the learned policy may be influenced by robot-infeasible motions.
Consequently, the learned policy may be encouraged to reproduce robot-infeasible motions, potentially resulting in falls or degraded task performance.

\begin{figure}[t]
    \centering
    \includegraphics[width=\linewidth]{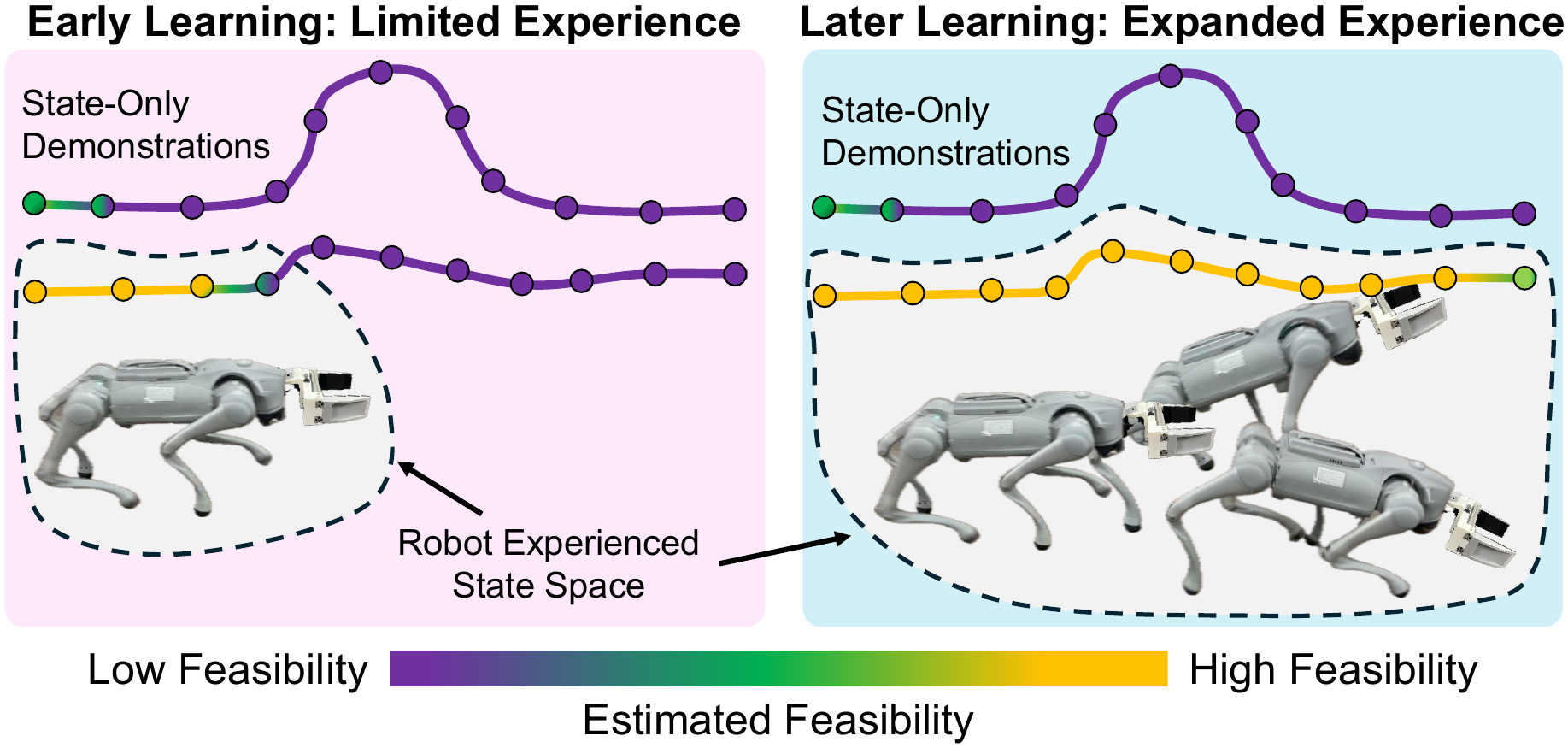}
    \caption{
    % Experience-based estimation for reducing the influence of infeasible demonstrations.
    Evolution of feasibility estimation during policy learning. Early in learning, limited robot experience yields high feasibility estimates only for demonstration transitions near the experienced region. As experience accumulates, the covered region expands, allowing more demonstration transitions to receive high feasibility estimates. This progressively reduces the influence of infeasible demonstrations during imitation learning.
    }
    \label{fig:motivation}
\end{figure}

To reduce the influence of infeasible demonstrations, 
model-based approaches evaluate whether demonstrated motions are feasible for the robot using a robot dynamics model \cite{takahashiICRA2025, chen2025}.
These methods predict the feasibility of demonstrated motions using known or learned robot dynamics models.
However, whole-body dynamics are strongly coupled, and contact interactions and balance constraints must also be considered.
Therefore, constructing an accurate dynamics model is not straightforward.

These limitations motivate a different approach to feasibility estimation: rather than explicitly modeling, \textit{what the robot can do, we infer feasibility from what it has actually experienced} (Fig. \ref{fig:motivation}).
During policy learning, the robot continuously generates state transitions that are physically realizable under its embodiment, providing direct evidence of feasible motion.
We therefore use this experience distribution as an embodiment-specific reference: demonstration transitions well supported by the robot's experience are likely to be feasible, without requiring an explicit dynamics model. % or relying on policy or discriminator outputs.

In this study, we propose Experience-based Feasibility-aware Generative Adversarial Imitation from Observation (EF-GAIfO).
EF-GAIfO estimates the feasibility of state-only demonstration transitions using experience data collected during GAIfO-based policy learning and reduces the influence of demonstrations with low estimated feasibility.
For feasibility estimation, we train a state-transition reconstruction model using the robot's experience data and evaluate demonstration transitions based on their reconstruction error.
The estimated feasibility is incorporated into the discriminator objective as a weight, allowing the policy to preferentially imitate demonstration transitions that are supported by the robot's experience.
We evaluate the proposed method through experiments in a point-mass environment and a quadruped robot environment.

%%%%%%%%%%%%%%%%%%%%%%%%%%%%%%%%%%%%%%%%%%%%%%%%%%%%%%%%%%%%%%%%%%%%%%%%%%%%%%%
%%%%%%%%%%%%%%%%%%%%%%%%%%%%%%%%%%%%%%%%%%%%%%%%%%%%%%%%%%%%%%%%%%%%%%%%%%%%%%%
%%%%%%%%%%%%%%%%%%%%%%%%%%%%%%%%%%%%%%%%%%%%%%%%%%%%%%%%%%%%%%%%%%%%%%%%%%%%%%%
\section{Related Work}
%%%%%%%%%%%%%%%%%%%%%%%%%%%%%%%%%%%%%%%%%%%%%%%%%%%%%%%%%%%%%%%%%%%%%%%%%%%%%%%

\subsection{Imitation Learning from Observation}

Unlike behavior cloning, which uses state-action pairs, imitation from observation aims to learn from demonstrations consisting only of state sequences or visual observations.
A possible approach to learn, despite the absence of expert actions, consists of inferring the missing actions through a learned inverse dynamics model that maps state transitions to actions \cite{torabi2018bco}.
Another approach to infer the missing action involves learning a latent action space directly from state observations that will be later aligned to the true action space using a small number of environment interactions \cite{edwards2019imitating}.
Liu et al. \cite{liu2018imitation}, instead, translate a demonstrator's raw video observations into the imitator's own context, enabling imitation in scenarios where the demonstrator's viewpoint or environment configuration is not the same as that of the imitator.
A separate line of work extends adversarial imitation learning to observation-only settings by matching distributions over state transitions instead of state-action pairs \cite{torabi2018gaifo}, using a Wasserstein formulation to handle rough or partial demonstrations \cite{li2023agile}, an inverse dynamics disagreement objective \cite{yang2019imitation}, or off-policy reuse of past interactions to improve sample efficiency \cite{zhu2020off}.

Previous methods for learning from observation mainly focus on observation-only or partial demonstrations, but they do not fully address the feasibility problem caused by differences between the demonstrator and the robot. When the demonstrator and the robot have different physical capabilities, some observed transitions may be difficult or impossible for the robot to reproduce. In contrast, this work introduces an experience-based feasibility estimation mechanism to reduce the influence of robot-infeasible transitions during imitation learning.

\subsection{Imitation Learning from Imperfect and Infeasible Demonstrations}
In practical imitation learning, demonstration data are not always optimal, clean, or suitable for the robot. Demonstrations may contain suboptimal actions, mistakes, noisy trajectories, or motions that cannot be executed by the robot. To address imperfect demonstrations, several studies have investigated methods for estimating the quality or confidence of demonstration data.
A common approach is to downweight low-quality demonstrations by estimating a per-sample confidence or quality score, either by using a subset of explicit confidence labels \cite{wu2019imitation}, or by adding a negative-risk term to counteract bias toward frequently occurring state-action pairs \cite{liu2025pn}.
In offline settings, one can distinguish expert from non-expert data without environment interaction, using cooperative discriminator training \cite{xu2022discriminator} or importance-sampling-weighted behavior cloning \cite{li2023imitation}.
Another popular approach consists of ranking the demonstrations by quality to extrapolate reward functions that can surpass the best available demonstration \cite{brown2019extrapolating}, an idea later extended to incorporate the generator's own experience into the ranking process \cite{shi2024ranking}.
Wang et al. \cite{wang2021learning}, instead, learn per-demonstration weights directly within the adversarial imitation learning framework.

A paradigm more related to the proposed method, even though less explored, is feasibility-aware imitation learning, which considers whether demonstrated motions are executable by the imitator.
Existing work addresses mismatched dynamics between demonstrator and learner \cite{cao2022learning}, the question of whether to follow potentially non-reproducible behavior \cite{lee2020selective}, and the estimation of feasibility for demonstrations collected via hand-mounted interfaces, using either a robot dynamics model \cite{takahashiICRA2025} or multimodal visual and haptic feedback \cite{takahashi2026multimodal}.

These methods focus on demonstration quality, require confidence labels, rely on explicit or learned robot dynamics models, or use prior robot execution data.
In contrast, the proposed method estimates demonstration feasibility from state transitions collected online during policy learning.
This enables the policy to reduce the influence of robot-infeasible demonstrations without requiring confidence labels, an explicit dynamics model, or a separate prior exploration dataset.
Therefore, this work focuses on experience-based feasibility estimation for IfO with demonstrations that may contain robot-infeasible motions.

%%%%%%%%%%%%%%%%%%%%%%%%%%%%%%%%%%%%%%%%%%%%%%%%%%%%%%%%%%%%%%%%%%%%%%%%%%%%%%%
%%%%%%%%%%%%%%%%%%%%%%%%%%%%%%%%%%%%%%%%%%%%%%%%%%%%%%%%%%%%%%%%%%%%%%%%%%%%%%%
%%%%%%%%%%%%%%%%%%%%%%%%%%%%%%%%%%%%%%%%%%%%%%%%%%%%%%%%%%%%%%%%%%%%%%%%%%%%%%%
\section{Generative Adversarial Imitation Learning from Observation (GAIfO)}
The objective of GAIfO \cite{torabi2018gaifo, torabi2019} is to learn a policy that can reproduce the state-only demonstrations $\mathcal D_\mathrm{demo}$ using reinforcement learning.
GAIfO assumes that the environment is modeled as a Markov decision process (MDP),
defined by the tuple $\mathcal{M} = (\mathcal{S}, \mathcal{A}, P, \rho_0, \gamma)$,
where $\mathcal{S}$ and $\mathcal{A}$ denote the state and action spaces,
$P(s_{t+1}\mid s_t,a_t)$ is the transition probability,
$\rho_0$ is the initial-state distribution,
and $\gamma \in [0,1)$ is the discount factor.

In GAIfO, the policy is trained using a discriminator $D$ to estimate similarity between its behavior and the demonstration data.
The discriminator $D$ is trained to distinguish between state transitions generated by the policy and those from expert demonstrations.
The discriminator is trained using demonstration data and exploration data collected by the policy $\pi$ as follows:
\begin{align}
    \max_{D}~
    &\mathbb E_{s_t, s_{t+1}\sim \mathcal D_\mathrm{demo}}\left[ \log D(s_t, s_{t+1}) \right] \nonumber \\
    + &\mathbb E_{\pi}\left[ \log \left(1 - D(s_t, s_{t+1})\right) \right].
\end{align}

The policy $\pi$ is then optimized using a reinforcement learning algorithm, such as PPO \cite{schulman2017proximal}, with rewards derived from the discriminator output:
\begin{align}
    r_t = -\log(1 - D(s_t, s_{t+1})).
    \label{eq:discriminator_reward}
\end{align}

%%%%%%%%%%%%%%%%%%%%%%%%%%%%%%%%%%%%%%%%%%%%%%%%%%%%%%%%%%%%%%%%%%%%%%%%%%%%%%%
%%%%%%%%%%%%%%%%%%%%%%%%%%%%%%%%%%%%%%%%%%%%%%%%%%%%%%%%%%%%%%%%%%%%%%%%%%%%%%%
%%%%%%%%%%%%%%%%%%%%%%%%%%%%%%%%%%%%%%%%%%%%%%%%%%%%%%%%%%%%%%%%%%%%%%%%%%%%%%%

\section{Experience-based Feasibility-aware GAIfO}
\begin{figure}[!t]
    \centering
    \includegraphics[width=0.95\linewidth]{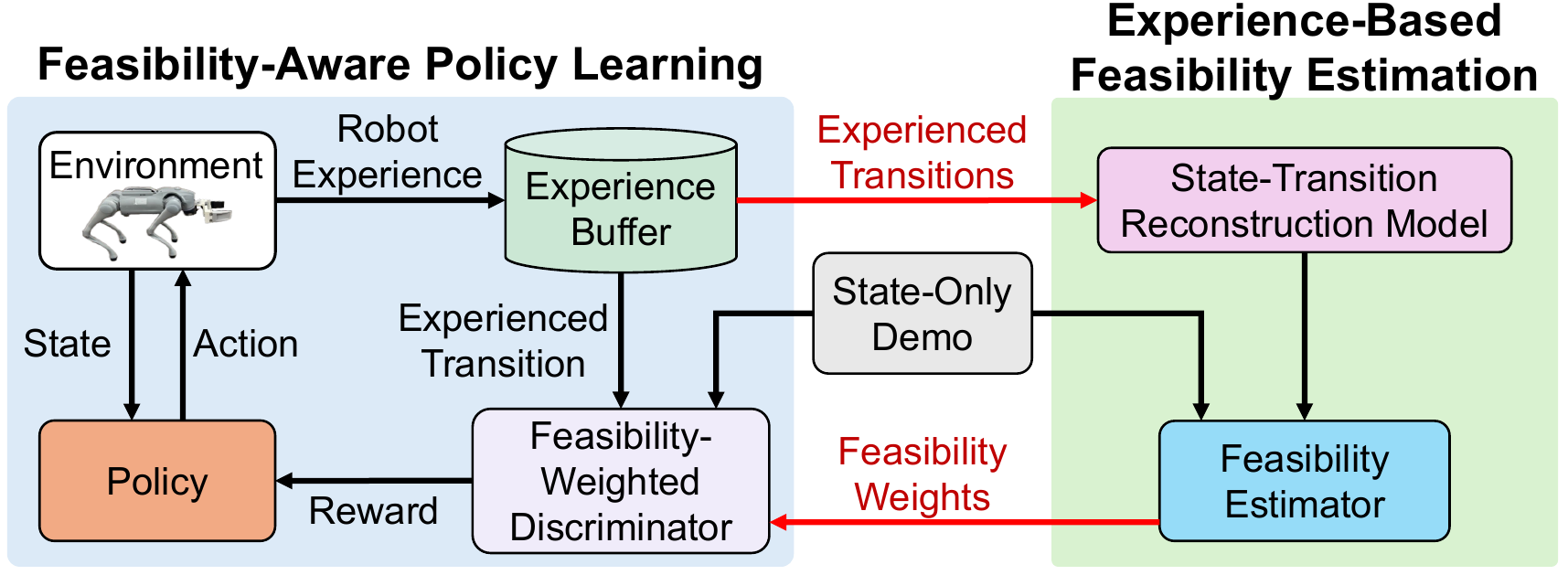}
    % \caption{Overview of EF-GAIfO}
    \caption{
        Overview of EF-GAIfO.
        Robot experience is used to estimate the feasibility of state-only demonstration transitions, and the estimated feasibility weights are incorporated into discriminator training for feasibility-aware policy learning.
    }
    \label{fig:EFGAIfO}
\end{figure}

This section describes Experience-based Feasibility-aware GAIfO (EF-GAIfO), an imitation learning framework for learning from demonstrations that may contain robot-infeasible motions. 
The overview of EF-GAIfO is shown in Fig. \ref{fig:EFGAIfO}.
First, we introduce an experience-based feasibility estimation method that evaluates state transitions in demonstrations by their similarity to those experienced by the robot.
Next, we incorporate the estimated feasibility into the GAIfO objective to reduce the influence of infeasible demonstrations during discriminator training.
Finally, we describe the overall learning procedure that jointly updates the policy, feasibility estimator, and discriminator.

%%%%%%%%%%%%%%%%%%%%%%%%%%%%%%%%%%%%%%%%%%%%%%%%%%%%%%%%%%%%%%%%%%%%%%%%%%%%%%%
%%%%%%%%%%%%%%%%%%%%%%%%%%%%%%%%%%%%%%%%%%%%%%%%%%%%%%%%%%%%%%%%%%%%%%%%%%%%%%%
\subsection{Feasibility Estimation}
Our feasibility estimation is based on the assumption that feasible demonstrations are similar to the state transitions experienced by the robot.
To estimate feasibility, our method learns a state-transition reconstruction model using the robot's experience data in GAIfO.
The feasibility of each demonstration is then evaluated based on the reconstruction error produced by the model.

The reconstruction model learns to reconstruct state-transition sequences generated by the robot during policy learning.
Given a transition of states $z_t=(s_t, s_{t+1})$, the reconstruction model $R$ produces the reconstructed sequence as $\hat{z_t} = R(z_t)$,
where, $\hat{z_t}$ is predicted state transition.

The reconstruction model is trained by minimizing the reconstruction loss $\mathcal L_R$ as:
\begin{align}
    \mathcal L_R = \sum_{t=1}^{T-1} \mid\mid z_t - R(z_t)\mid\mid^2,
    \label{eq:reconstruction_loss}
\end{align}
where $T$ denotes the length of the state sequence.

The feasibility of demonstrated motions is evaluated using the reconstruction model $R$ based on reconstruction error.
For state transition $z_t$, the error $e(z_t, R(z_t))$ is calculated:
\begin{align}
    e(z_t, R(z_t)) =
    \frac{1}{D_s}
    \sum_{d=1}^{D_s}
    \max_{h \in \{0,1\}}
    \left(
        s_{t+h,d} - \hat{s}_{t+h,d}
    \right)^2 ,
    \label{eq:reconstruction_error}
\end{align}
where $D_s$ is the number of state dimensions, 
and $h$ means time step in a state transition,
and $s_{t,d}$ and $\hat{s}_{t,d}$ denote the original and reconstructed state, respectively, at state dimension $d$.
We use the maximum error over the two consecutive states so that a transition receives a large error if either state is poorly reconstructed.

The feasibility is computed from the error $e$ as a value between 0 and 1:
\begin{align}
    f_w(z_t, R(z_t)) =
    \operatorname{clip}
    \left(
    \frac{\tau_{\mathrm{high}}-e(z_t, R(z_t))}
         {\tau_{\mathrm{high}}-\tau_{\mathrm{low}}},
    0,1
    \right),
    \label{eq:feasibility_weight}
\end{align}
where, 
$\operatorname{clip}(\cdot, 0, 1)$ denotes a clipping function that restricts the output value to the range $[0,1]$,
$\tau_{\mathrm{low}}$ and $\tau_{\mathrm{high}}$ are the lower and upper thresholds, respectively.

%%%%%%%%%%%%%%%%%%%%%%%%%%%%%%%%%%%%%%%%%%%%%%%%%%%%%%%%%%%%%%%%%%%%%%%%%%%%%%%
%%%%%%%%%%%%%%%%%%%%%%%%%%%%%%%%%%%%%%%%%%%%%%%%%%%%%%%%%%%%%%%%%%%%%%%%%%%%%%%
\subsection{Feasibility-weighted Objective}
The objective of EF-GAIfO uses the estimated feasibility weights to control the contribution of each demonstrated motion to the discriminator training:
\begin{align}
    \max_{D}~ &\mathbb{E}_{\pi_e}
    \left[
    f_w(z_t, R(z_t))
    \log D(s_t,s_{t+1})
    \right] \nonumber \\ 
    + &\mathbb{E}_{\pi}
    \left[
    \log\left(1-D(s_t,s_{t+1})\right)
    \right].
    \label{eq:feasibility_aware_gail}
\end{align}
By incorporating a feasibility weight into the discriminator's objective for demonstration data, the policy is trained to prioritize demonstrations with high feasibility.
The feasibility weight $f_w$ is calculated using Eq.~\ref{eq:feasibility_weight}, which reduces the effect of expert transitions with low estimated feasibility.

\subsection{Learning Process}

\begin{algorithm}[t]
    \caption{Learning Process of EF-GAIfO}
    \label{alg:fgail}
    % \scriptsize
    % \footnotesize
    \algrenewcommand\algorithmicindent{0.7em}
    \begin{algorithmic}[1]
    \State \textbf{Input:} dataset of state-only demonstration data $\mathcal{D}$
    \State \textbf{Initialize:} discriminator $D$, reconstruction model $R$,
    policy $\pi$, value function $V$, and experience buffer $\mathcal{B}_{\pi}$.
    \For{learning iteration $i=1,2,\ldots, I$ }
        \State Collect $N+H$ experienced transitions $(s_t,a_t,s_{t+1})_{t-H}^{t+N}$ using $\pi$
        \State Compute rewards according to Eq.~\eqref{eq:discriminator_reward}

        \State Sample mini-batch $b^{\pi}\sim\mathcal{B}_{\pi}$
        \State Update policy $\pi$ and value func. $V$ using PPO with mini-batch $b^{\pi}$
        \State Update reconstruction model $R$ using $b^{\pi}$ by Eq.~\eqref{eq:reconstruction_loss}
        
        \State Sample mini-batch $b^{E}\sim\mathcal{D}$
        \State Compute reconstruction error $e$ for $b^{E}$
        % \State Update parameters $\tau_{\mathrm{low}}$ and $\tau_{\mathrm{high}}$
        \State Compute feasibility weight $f_w$ by Eq.~\eqref{eq:feasibility_weight}
        \State Update discriminator $D$ using $b^{\pi}$, $b^{E}$ and $f_w$ by  Eq.~\eqref{eq:feasibility_aware_gail}
    \EndFor
    \end{algorithmic}
\end{algorithm}

The EF-GAIfO algorithm updates the reconstruction model, the policy, and the discriminator.
The feasibility weights computed using the reconstruction model are then used to update the discriminator.
Algorithm~\ref{alg:fgail} summarizes the learning process of EF-GAIfO.
At each iteration, the current policy collects experience transitions and is updated using rewards (lines 3--7).
The collected experience is also used to update the reconstruction model (line 8).
The reconstruction model then evaluates demonstration samples and computes their feasibility weights (lines 9--11).
Finally, the discriminator is updated using the feasibility-weighted demonstrations and the robot experience (line 12).

The thresholds $\tau_\mathrm{high}$ and $\tau_\mathrm{low}$ for the feasibility weight are updated based on the reconstruction errors of the demonstration samples, with the percentile parameter $q$ gradually increased during training.

%%%%%%%%%%%%%%%%%%%%%%%%%%%%%%%%%%%%%%%%%%%%%%%%%%%%%%%%%%%%%%%%%%%%%%%%%%%%%%%
%%%%%%%%%%%%%%%%%%%%%%%%%%%%%%%%%%%%%%%%%%%%%%%%%%%%%%%%%%%%%%%%%%%%%%%%%%%%%%%
%%%%%%%%%%%%%%%%%%%%%%%%%%%%%%%%%%%%%%%%%%%%%%%%%%%%%%%%%%%%%%%%%%%%%%%%%%%%%%%
\section{Experiments}
In this section, we evaluate EF-GAIfO using both controlled point-mass experiments and a quadruped robot task with human-collected demonstrations.
The research questions addressed in our experiments are as follows:
\begin{enumerate}
    \item[\textbf{RQ1:}] Can experience-based feasibility estimation distinguish feasible from infeasible demonstration transitions as robot experience accumulates?
    \item[\textbf{RQ2:}] Does experience-based feasibility weighting improve imitation performance across different forms of feasibility mismatch?
    \item[\textbf{RQ3:}] Can EF-GAIfO learn effectively from human-collected state-only demonstrations and enable successful execution on a real robot?
\end{enumerate}
We evaluate RQ1 and RQ2 in a point-mass environment in Sec. V.A, and RQ3 in a quadruped robot environment in Sec. V.B.

\subsection{Point-Mass Environment}
\label{subsec:point_mass_environment}

%%%%%%%%%%%%%%%%%%%%%%%%%%%%%%%%%%%%%%%%%%%%%%%%%%%%

\subsubsection{Settings}
\label{subsubsec:point_mass_settings}

We first evaluate EF-GAIfO in a two-dimensional point-mass environment to address RQ1 and RQ2 under controlled conditions.
Specifically, we examine how the distinction between feasible and infeasible demonstration transitions evolves as robot experience accumulates and whether experience-based feasibility weighting improves imitation performance across different forms of feasibility mismatch.
In this environment, the feasibility of each demonstrated transition can be explicitly determined from the agent's velocity limits.

The key design of this experiment is to separate two different types of feasibility selection.
The multi-speed setting evaluates whether EF-GAIfO can select feasible motion speeds when the demonstrated path is fixed.
In contrast, the multi-path setting evaluates whether EF-GAIfO can selectively imitate feasible trajectory modes when feasible and infeasible paths are mixed in the demonstration dataset.
Thus, the two settings test complementary capabilities:
feasibility discrimination along the same path and selective imitation among multiple trajectory modes.

The objective of the task is for an agent to move from a fixed start position at \((0,0)\) m to a goal position at \((8,0)\) m.
The state is defined as the two-dimensional position \((x, y)\), and the action is defined as a two-dimensional velocity command.
Each episode contains a maximum of 30 steps.
% An episode is considered unsuccessful when the policy exceeds the point-mass velocity limit.
An episode terminates when the maximum episode length is reached or when the policy exceeds the point-mass velocity limit. A trial is considered successful if the point-mass reaches the goal position before termination.
The velocity limits $(v_x, v_y)$ are set to $(\pm3.9,\,\pm2.5)$ m/s and $(\pm4.7,\,\pm2.5)$ m/s for the multi-speed and multi-path settings, respectively.

\begin{figure}[t]
    \centering
    \begin{minipage}[b]{\linewidth}
        \centering
        \includegraphics[width=0.9\linewidth]{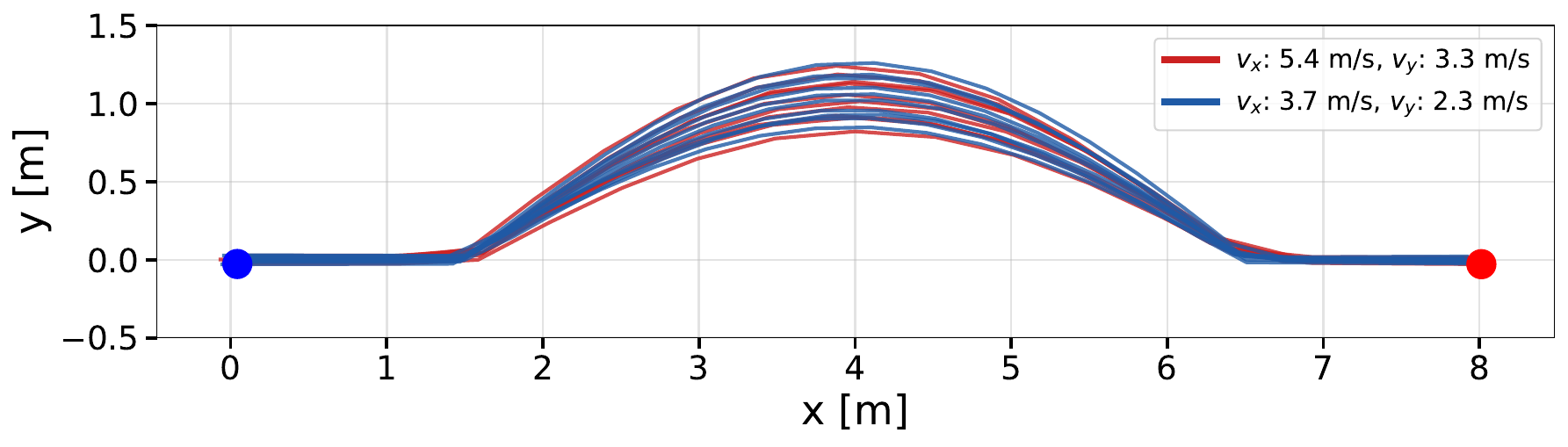}
        \vspace{-2mm}
        \subcaption{Multi-speed demonstrations}
        \label{fig:point_mass_demonstrations_multi_speed}
    \end{minipage}
    \begin{minipage}[b]{\linewidth}
        \centering
        \includegraphics[width=0.9\linewidth]{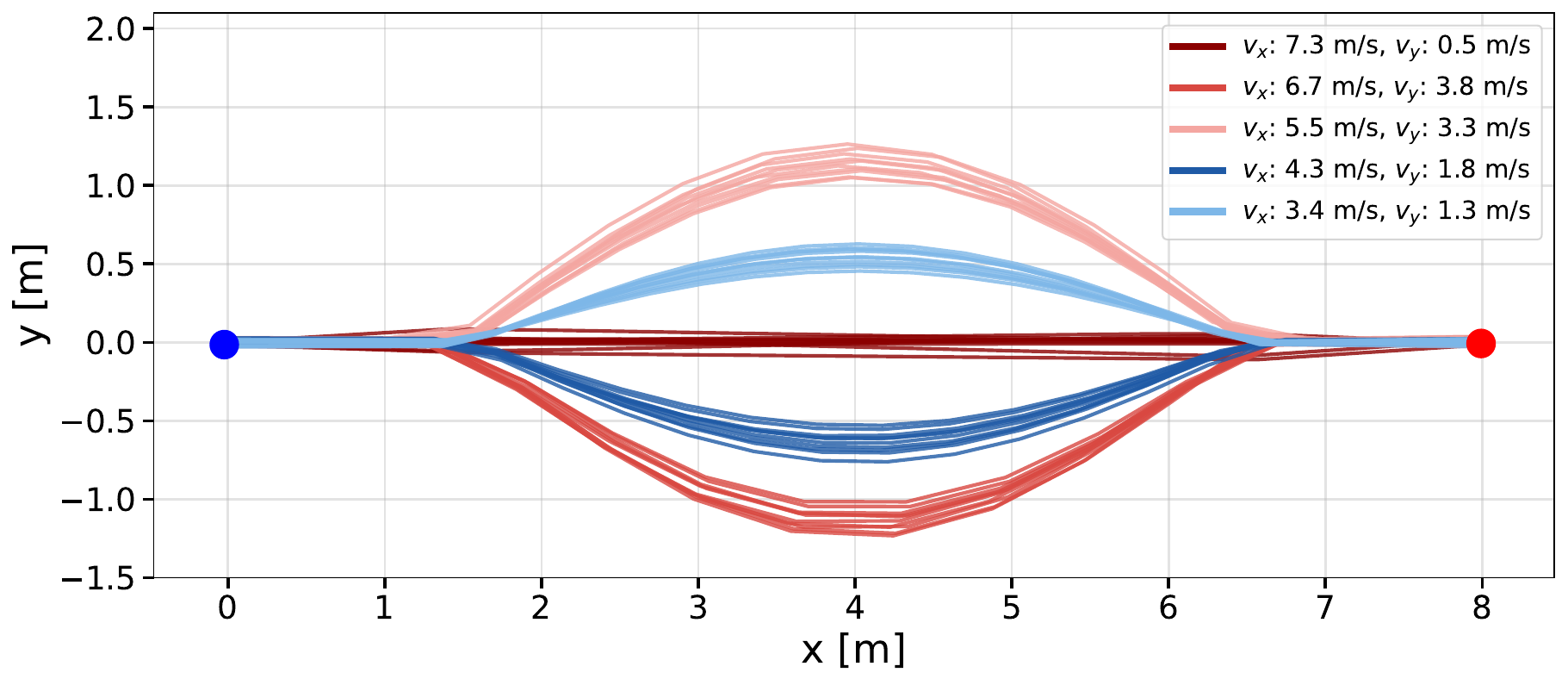}
        \vspace{-2mm}
        \subcaption{Multi-path demonstrations}
        \label{fig:point_mass_demonstrations_multi_paths}
    \end{minipage}
    % \caption{Demonstration trajectories in the multi-path demonstration experiment.}
    \caption{
        % Demonstration trajectories in the point-mass environment.
        % The labels indicate the maximum absolute velocities \((|v_x|_{\max}, |v_y|_{\max})\) in m/s for each path.
        Demonstration trajectories in the point-mass environment for (a) the multi-speed setting and (b) the multi-path setting.
        Red and blue trajectories represent infeasible and feasible demonstrations, respectively.
        Color intensity indicates execution speed, with lighter colors denoting slower motions.
    }
    \label{fig:point_mass_demonstrations}
\end{figure}

In the multi-speed setting, we use two types of demonstrations that follow nearly identical paths but differ in execution speed.
We collect ten trajectories for each speed, resulting in 20 demonstration trajectories in total.
The slower demonstrations satisfy the agent's velocity limits, whereas the faster demonstrations exceed them.
The demonstration trajectories and their velocities are shown in Fig.~\ref{fig:point_mass_demonstrations_multi_speed}.

In the multi-path demonstration, five demonstration paths with varying trajectories and execution speeds are collected.
Ten demonstrations are collected for each path, resulting in a total of 50 demonstration trajectories.
Regarding the agent's speed limits, we configure the settings so that two of the five demonstration paths are feasible.
The demonstration trajectories and their speeds are shown in Fig. \ref{fig:point_mass_demonstrations_multi_paths}.

For EF-GAIfO, the policy is trained for 3000 iterations, with 30 interaction steps collected from each environment per iteration.
The actor, critic, and discriminator networks are each implemented as a multilayer perceptron with hidden layers of 512 and 256 units.
% The discriminator is also implemented as a multilayer perceptron with two hidden layers of 512 and 256 units.
The reconstruction model is implemented as a variational autoencoder \cite{kingma2014}.
Its encoder and decoder use two hidden layers of 64 and 32 units, and the latent dimension is set to two.
All hidden layers use ReLU activation functions, and all networks are optimized using Adam.
We compare EF-GAIfO with GAIfO and WGAIfO in both experiments.
WGAIfO is an adaptation of WGAIL \cite{wang2021learning} to the GAIfO setting, where the demonstration weighting mechanism originally proposed for identifying imperfect demonstrations is applied to state-only imitation learning.
Each method is trained with 20 random seeds and evaluated over 10 episodes per seed, resulting in 200 trials for each setting.

\subsubsection{Results}
\label{subsubsec:point_mass_results}

\begin{table}[!t]
    \centering
    \setlength{\tabcolsep}{4pt}
    \caption{
    Mean task success rates and Hausdorff distances in the point-mass experiment.
    The Hausdorff distance measures the discrepancy between the trajectories generated by the learned policy and the feasible demonstration trajectories.
    Values are reported as the mean $\pm$ standard deviation of the learned policy over 20 random seeds.
    Each learned policy is evaluated over ten episodes.
    }
    \label{tab:point_mass_result_success_rate}
    \renewcommand{\arraystretch}{1.15}
    \begin{tabular}{cccc}
        \toprule
        Setting     & Method & Success rate ($\uparrow$) & Hausdorff distance ($\downarrow$)\\
        \midrule
        \midrule
                    & EF-GAIfO & \(\mathbf{90.0 \pm 30.0\%}\)  & \(\mathbf{0.884 \pm 1.72}\) \\
        Multi-speed & WGAIfO   & \(20.0 \pm 44.7\%\)           & \(2.629 \pm 2.694\) \\
                    & GAIfO    & \(15.0 \pm 35.7\%\)           & \(4.877 \pm 2.396\) \\
        \midrule                                       
                    & EF-GAIfO & \(\mathbf{100.0 \pm 0.0\%}\)  & \(\mathbf{0.163 \pm 0.075}\) \\
        Multi-path  & WGAIfO   & \(20.0 \pm 41.0\%\)           & \(1.742 \pm 2.343\) \\
                    & GAIfO    & \(50.0 \pm 50.0\%\)           & \(2.828 \pm 2.57\) \\
        \bottomrule
    \end{tabular}
\end{table}

\begin{table}[!t]
    \centering
    \setlength{\tabcolsep}{4pt}
    \caption{
        Mean evaluation values for feasible and infeasible demonstration data across training iterations, assigned using EF-GAIfO and WGAIfO trained with 20 different random seeds in the point-mass experiments.
        The evaluation values should be high for feasible demonstrations and low for infeasible demonstrations.
    }
    \label{tab:point_result_mean_weights}
    \begin{tabular}{ccccc}
        \toprule
        Setting       & Demonstration            & Iter. & EF-GAIfO            & WGAIfO \\
        \midrule
        \midrule
        % \multirow{4}{*}{Multi-speed}
                    & Feasible ($\uparrow$)      & 100  & 0.142 $\pm$ 0.032 & 0.450 $\pm$ 0.039 \\
        Multi-      & Feasible ($\uparrow$)      & 3000 & 0.924  $\pm$ 0.105  & 0.402 $\pm$ 0.473 \\
        speed       & Infeasible ($\downarrow$)  & 100  & 0.016 $\pm$ 0.030 & 0.023 $\pm$ 0.005 \\
                    & Infeasible ($\downarrow$)  & 3000 & 0.141  $\pm$ 0.166  & 0.0001 $\pm$ 0.0004 \\
        \midrule  
        % \multirow{4}{*}{Multi-path}
                    & Feasible ($\uparrow$)      & 100  & 0.152  $\pm$ 0.031  & 0.301 $\pm$ 0.068 \\
        Multi-      & Feasible ($\uparrow$)      & 3000 & 0.721  $\pm$ 0.021  & 0.298 $\pm$ 0.030 \\
        path        & Infeasible ($\downarrow$)  & 100  & 0.0035 $\pm$ 0.0044 & 0.007 $\pm$ 0.012 \\
                    & Infeasible ($\downarrow$)  & 3000 & 0.0707 $\pm$ 0.0336 & 0.001 $\pm$ 0.003 \\
        \bottomrule
    \end{tabular}
\end{table}

%%%%%%%%%%%%%%%%%%%%%%%%%%%%%%%%%%%%%%%%%%%%%%%%%%%%%%%%%%%%%%%%%%%%%%
\begin{figure}[t]
    \centering
    % Feasibility weight colorbar
    \includegraphics[width=0.6\linewidth]{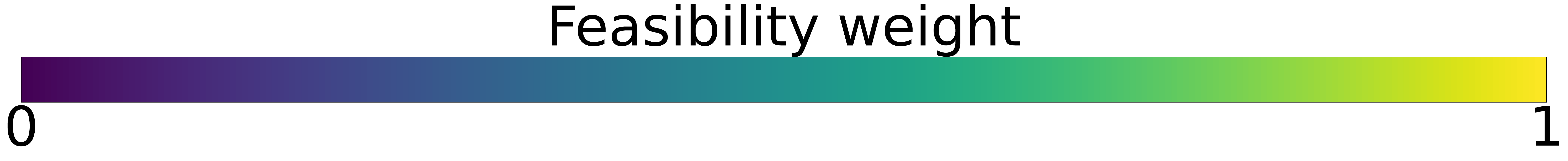}
    \begin{minipage}[t]{0.47\linewidth}
        \centering
        \includegraphics[width=\linewidth]{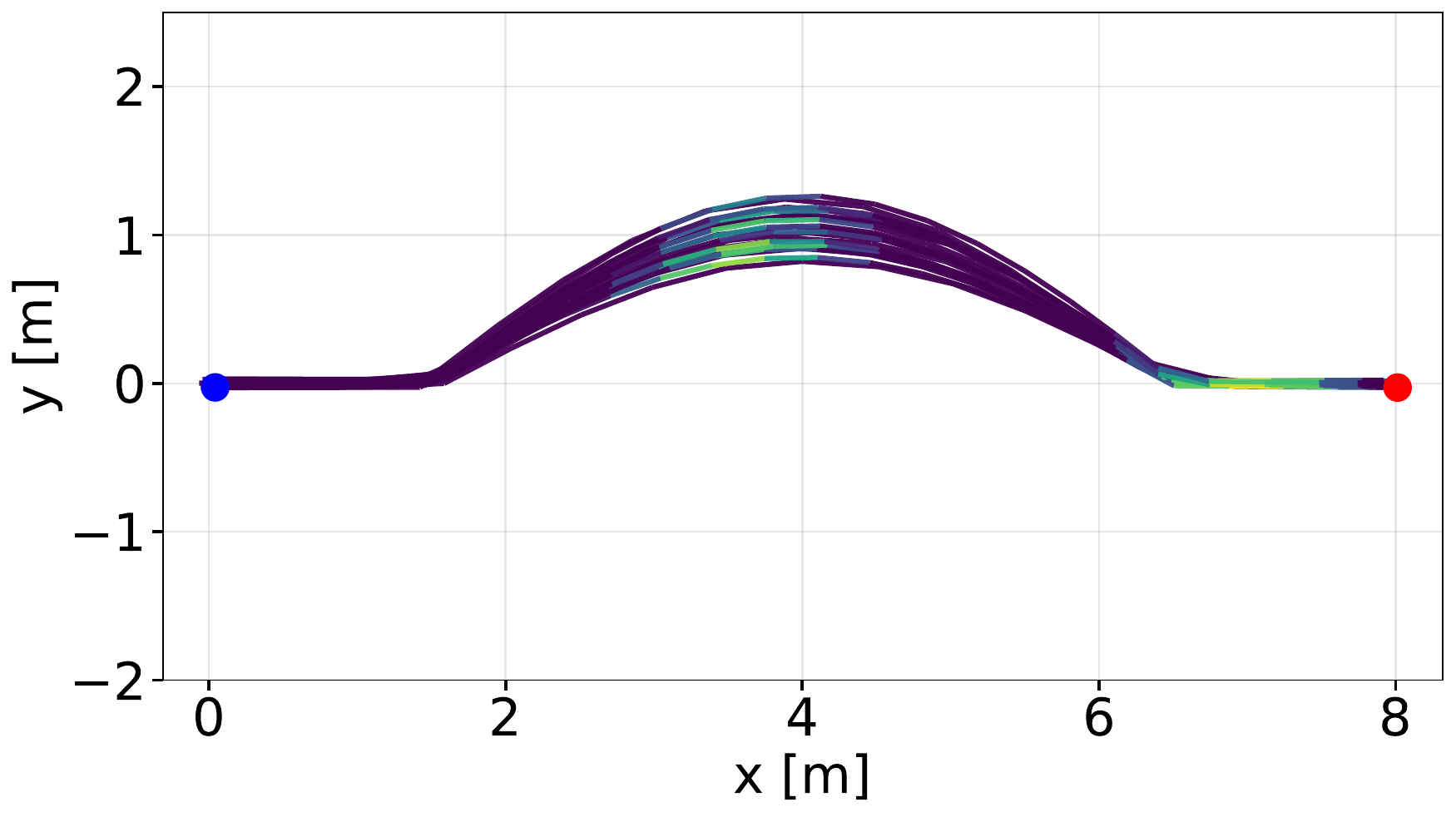}
        \vspace{-6mm}
        \subcaption{Multi-speed setting at 100 iter.}
        \label{fig:point_mass_feasibility_visualization:multi_speed:100}
    \end{minipage}
    \hfill
    \begin{minipage}[t]{0.47\linewidth}
        \centering
        \includegraphics[width=\linewidth]{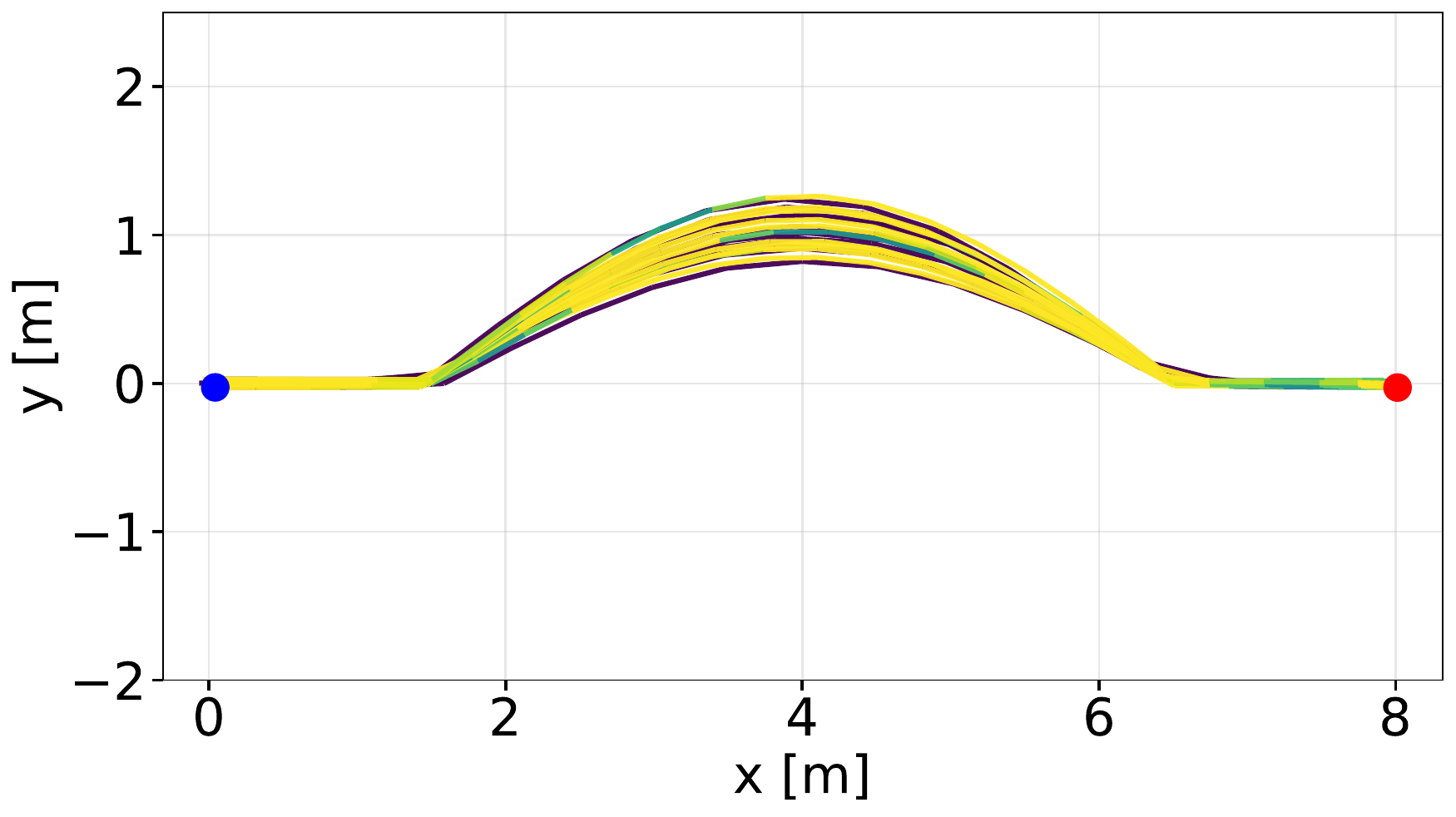}
        \vspace{-6mm}
        \subcaption{Multi-speed setting at 3000 iter.}
        \label{fig:point_mass_feasibility_visualization:multi_speed:2999}
    \end{minipage}
    \begin{minipage}[t]{0.48\linewidth}
        \centering
        \includegraphics[width=\linewidth]{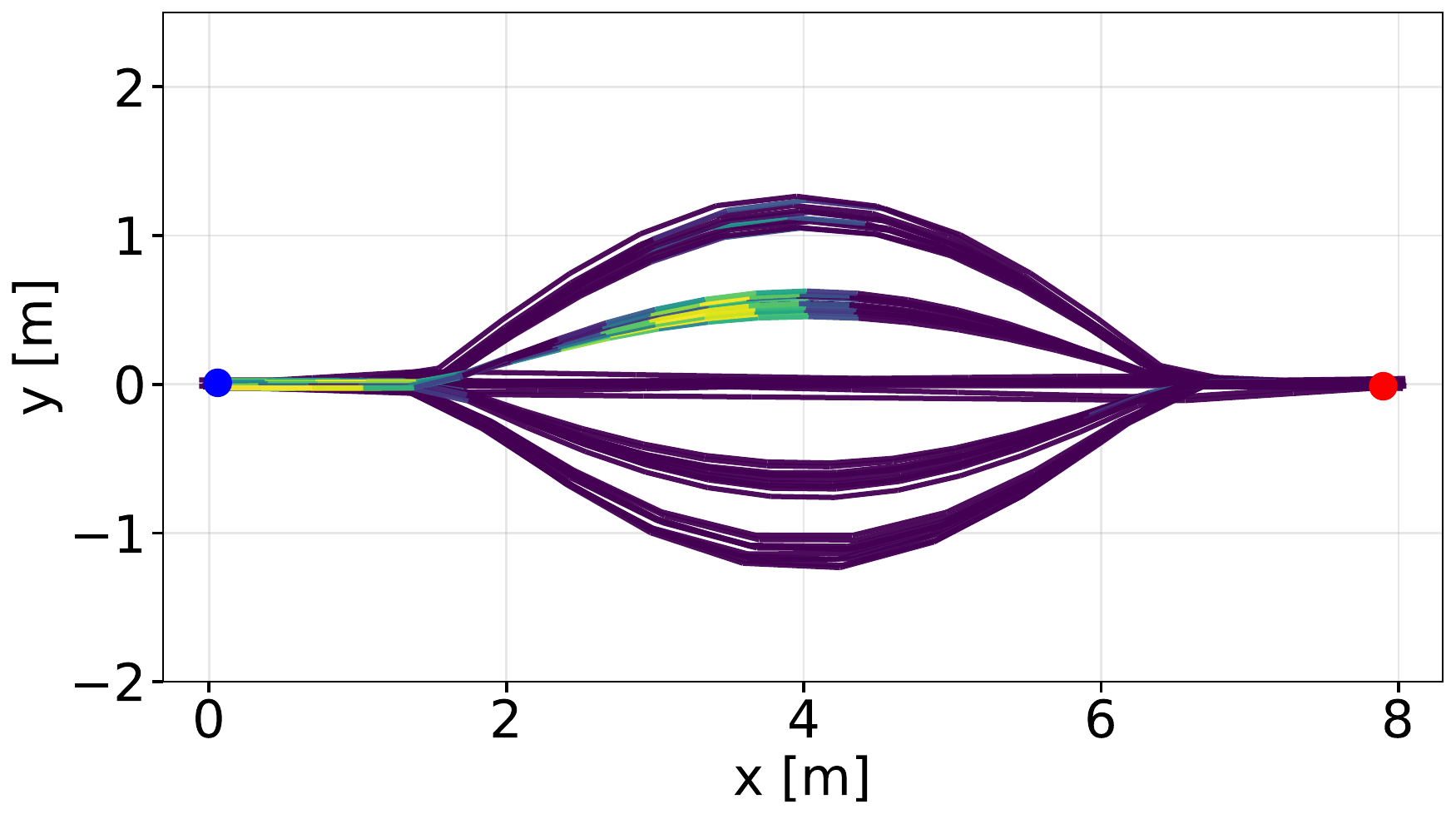}
        \vspace{-6mm}
        \subcaption{Multi-path setting at 100 iter.}
        \label{fig:point_mass_feasibility_visualization:multi_path:100}
    \end{minipage}
    \hfill
    \begin{minipage}[t]{0.48\linewidth}
        \centering
        \includegraphics[width=\linewidth]{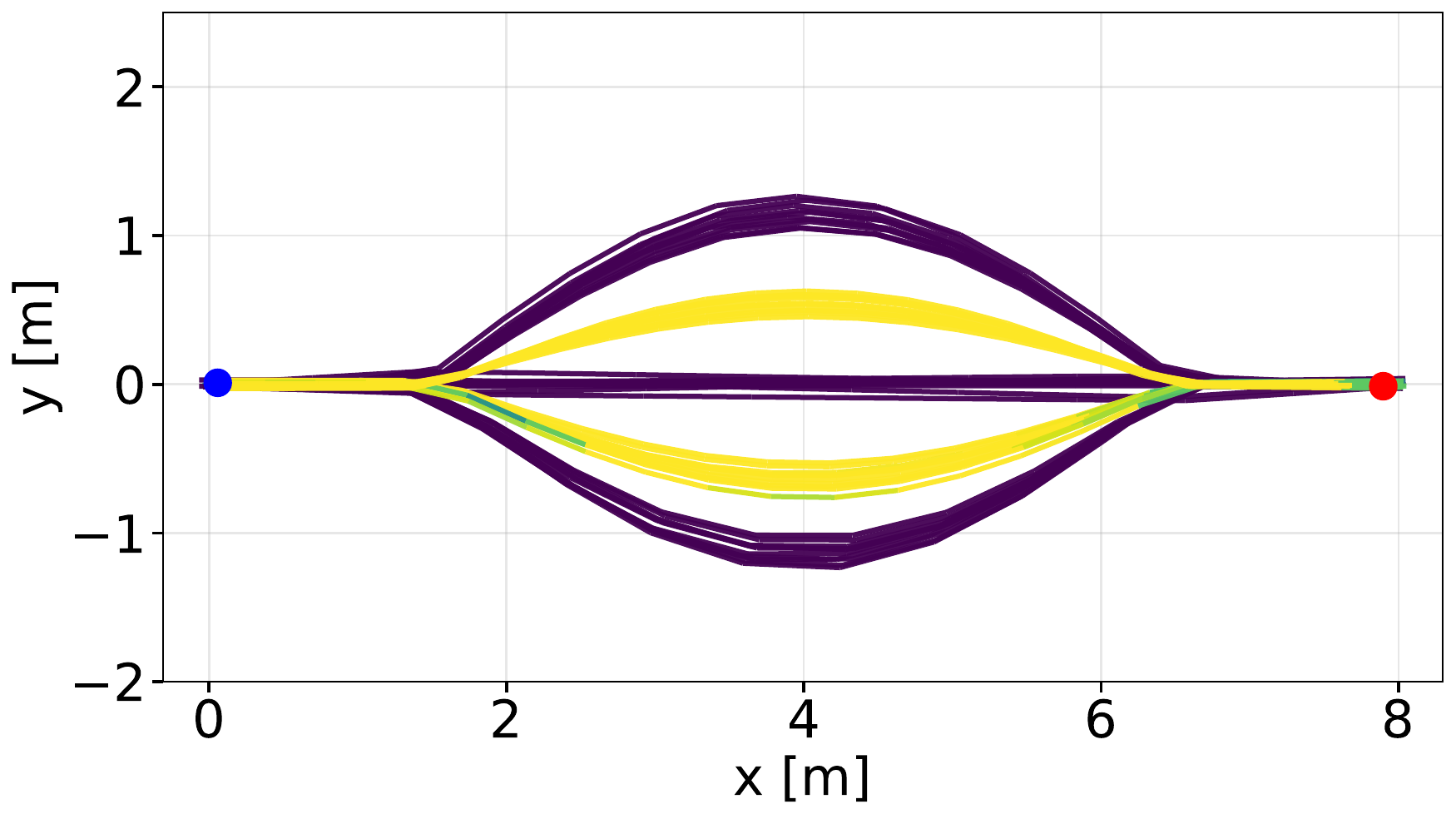}
        \vspace{-6mm}
        \subcaption{Multi-path setting at 3000 iter.}
        \label{fig:point_mass_feasibility_visualization:multi_path:2999}
    \end{minipage}
    \caption{
        Feasibility weights estimated by EF-GAIfO in the point-mass experiments. 
        (a) and (b) show the multi-speed setting at 100 and 3000 training iterations, respectively, while (c) and (d) show the multi-path setting. 
        The color of each demonstration trajectory represents its estimated feasibility weight.
        %, and the red trajectories indicate policy exploration trajectories.
    }
    \label{fig:point_mass_feasibility_visualization}
\end{figure}

%%%%%%%%%%%%%%%%%%%%%%%%%%%%%%%%%%%%%%%%%%%%%%%%

%%%%%%%%%%%%%%%%%%%%%%%%%%%%%%%%%%%%%%%%%%%%%%%%

Table~\ref{tab:point_mass_result_success_rate} summarizes the task success rates and Hausdorff distances for the point-mass experiments with multi-speed and multi-path settings.
In both the multi-speed and multi-path settings, EF-GAIfO achieved the highest success rate and the smallest Hausdorff distance, outperforming both WGAIfO and GAIfO.
The results indicate that EF-GAIfO can reduce the negative influence of the fast infeasible demonstrations and learn a policy that follows the feasible demonstrations more closely.
Specifically for the multi-path setting, these results suggest that EF-GAIfO can selectively imitate feasible trajectory modes.

Table~\ref{tab:point_result_mean_weights} reports the mean evaluation values assigned to feasible and infeasible demonstrations at different training iterations.
In both the multi-speed and multi-path settings, EF-GAIfO assigns relatively low values at 100 iterations due to limited policy experience.
As training progresses, the values for feasible demonstrations increase substantially, whereas those for infeasible demonstrations remain low.
In contrast, WGAIfO does not show the same progressive separation between feasible and infeasible demonstrations.
This trend is consistently observed in both experimental settings.

Figure~\ref{fig:point_mass_feasibility_visualization} visualizes the feasibility weights estimated by EF-GAIfO at 100 and 3000 training iterations.
At 100 iterations, high feasibility weights appear only in a limited region of the demonstrations, reflecting the limited policy experience in the early stage of training.
At 3000 iterations, high feasibility weights are assigned over a larger region of the feasible trajectories.
In contrast, infeasible demonstrations remain assigned relatively low weights.

Taken together, these results answer RQ1 and RQ2. Regarding RQ1, the feasibility estimates become more discriminative as the robot's experience accumulates: the weights assigned to feasible demonstration transitions increase substantially, whereas those assigned to infeasible transitions remain low. Regarding RQ2, EF-GAIfO achieves higher task success rates and smaller Hausdorff distances than both baselines in the multi-speed and multi-path settings. Thus, experience-based feasibility weighting improves imitation performance across different forms of feasibility mismatch.

%%%%%%%%%%%%%%%%%%%%%%%%%%%%%%%%%%%%%%%%%%%%%%%%%%%%%%%%%%%
\subsection{Quadruped Robot Environment}
\subsubsection{Settings}

\begin{figure}[t]
    \centering
    \includegraphics[width=0.7\linewidth]{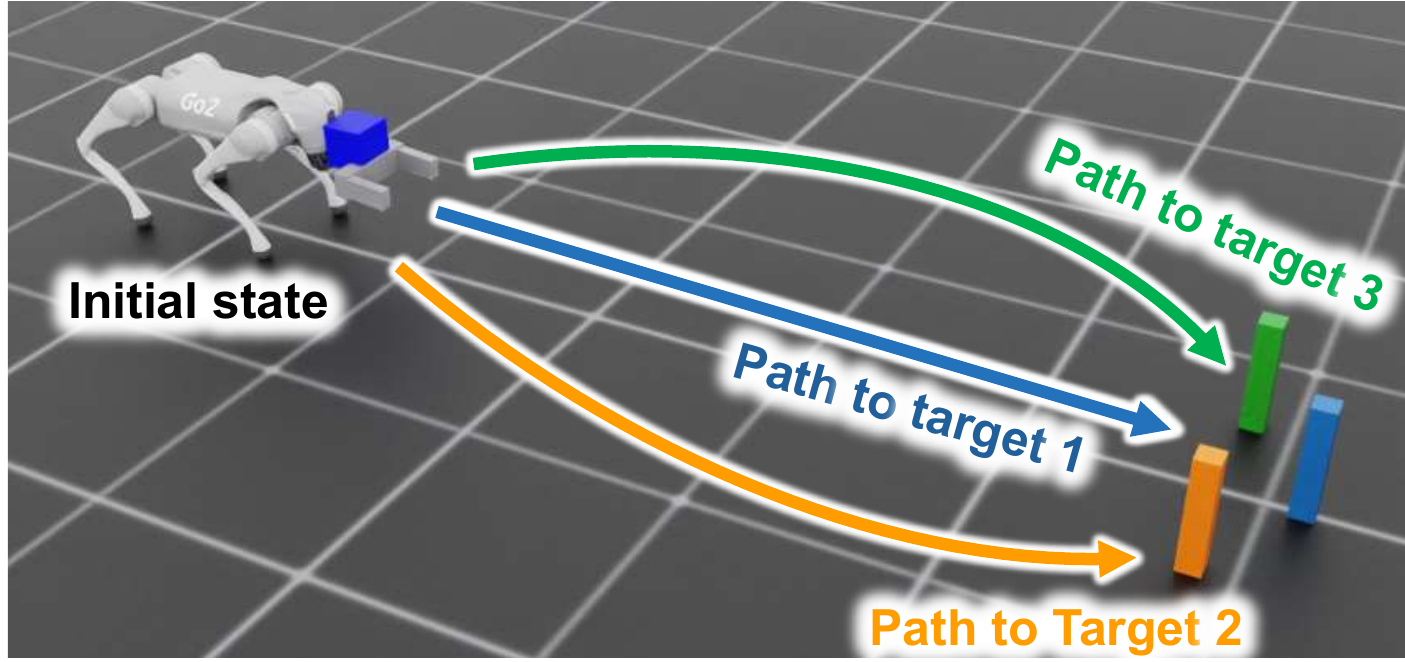}
    % \vspace{-2mm}
    \caption{
        Object-reaching-and-picking task setting in the quadruped robot environment.
    }
    \label{fig:go2_task_setting}
\end{figure}

\begin{figure}[!t]
    \centering
    \begin{minipage}[t]{0.48\linewidth}
        \centering
        \includegraphics[width=0.95\linewidth]{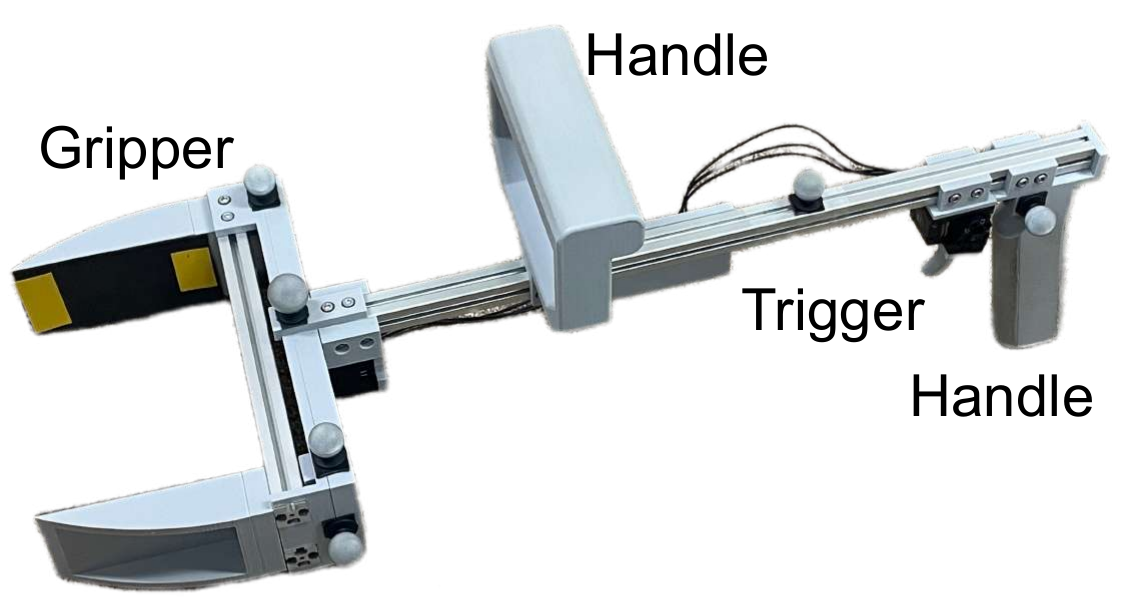}
        \subcaption{Robot-free demonstration interface}
        \label{fig:interface}
    \end{minipage}
    \begin{minipage}[t]{0.48\linewidth}
        \centering
        \includegraphics[width=0.95\linewidth]{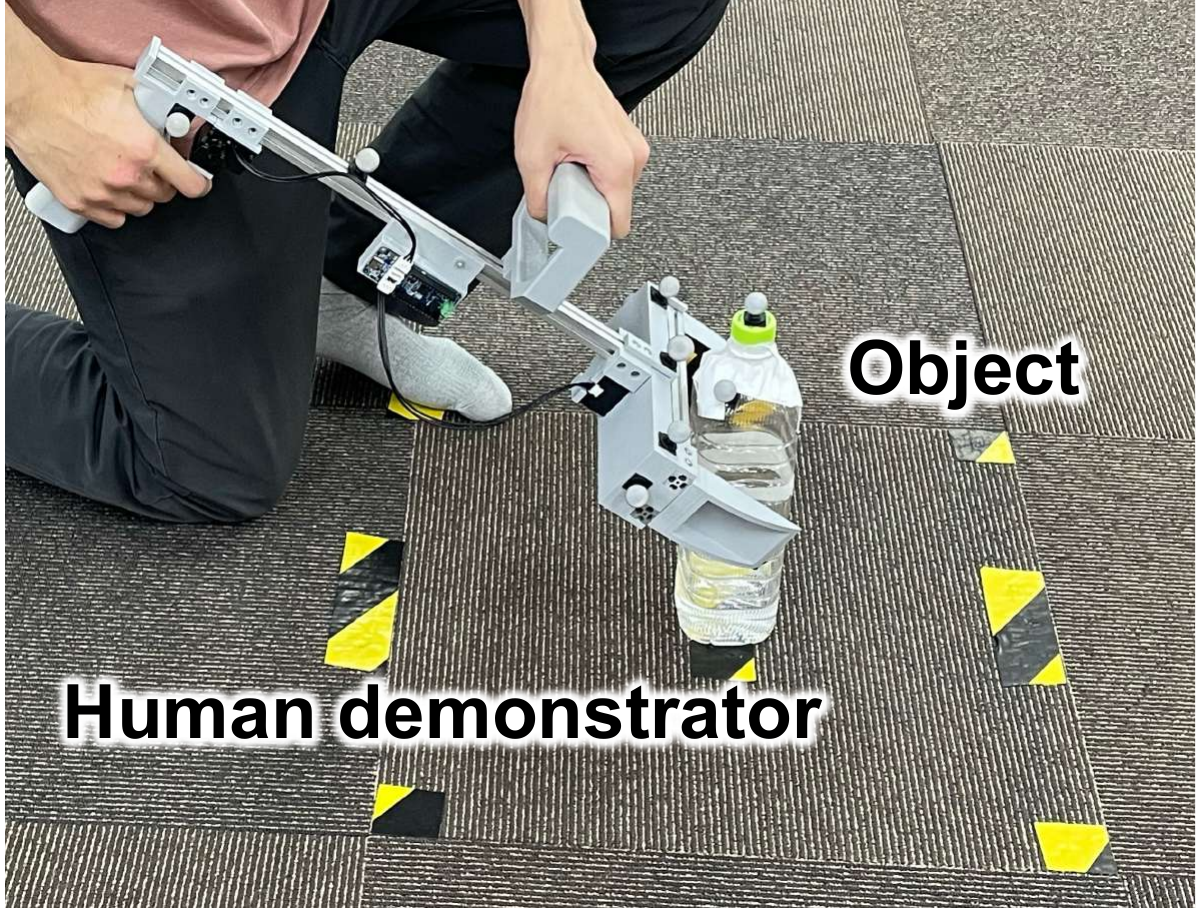}
        \subcaption{Demonstration for robot experiment}
        \label{fig:robot:experient:demo}
    \end{minipage}
    \caption{Demonstration for robot experiment}
    \label{fig:robot:experient:demo}
\end{figure}

%%%%%%%%%%%%%%%%%%%%%%%%%%%%%%%%%%%%%%%%%%%%%%%%%%%%%%%%%%%%%%
%human demo plot
\begin{figure}[t]
    \centering
    \includegraphics[width=0.7\linewidth]{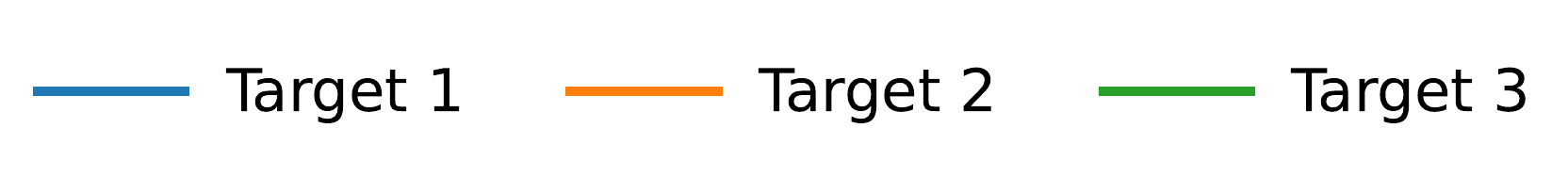}
    \vspace{1mm}
    \begin{subfigure}[t]{0.48\linewidth}
        \centering
        \includegraphics[width=\linewidth]{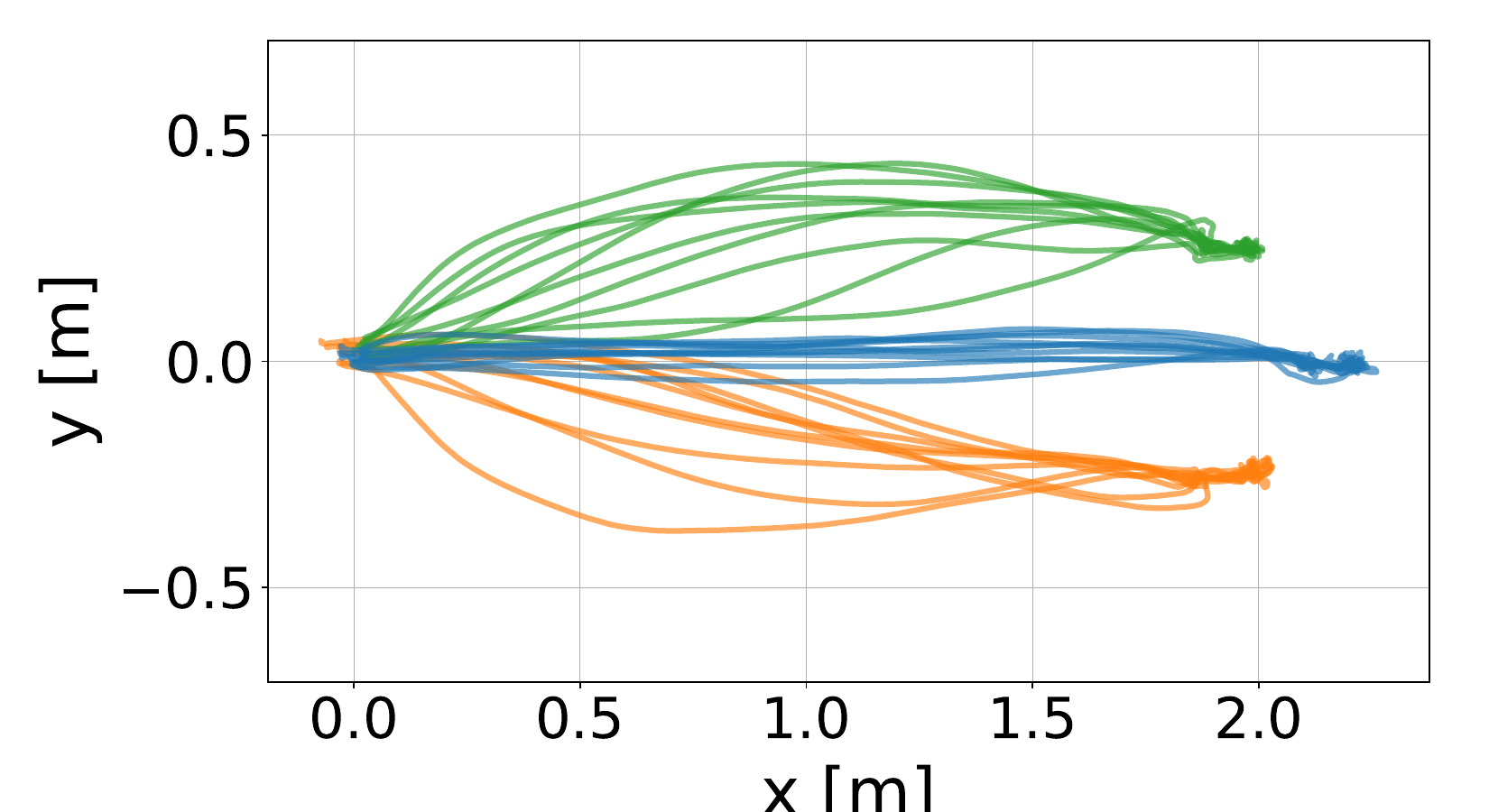}
        \subcaption{XY trajectories}
        \label{fig:human_demo_xy}
    \end{subfigure}
    \hfill
    \begin{subfigure}[t]{0.48\linewidth}
        \centering
        \includegraphics[width=\linewidth]{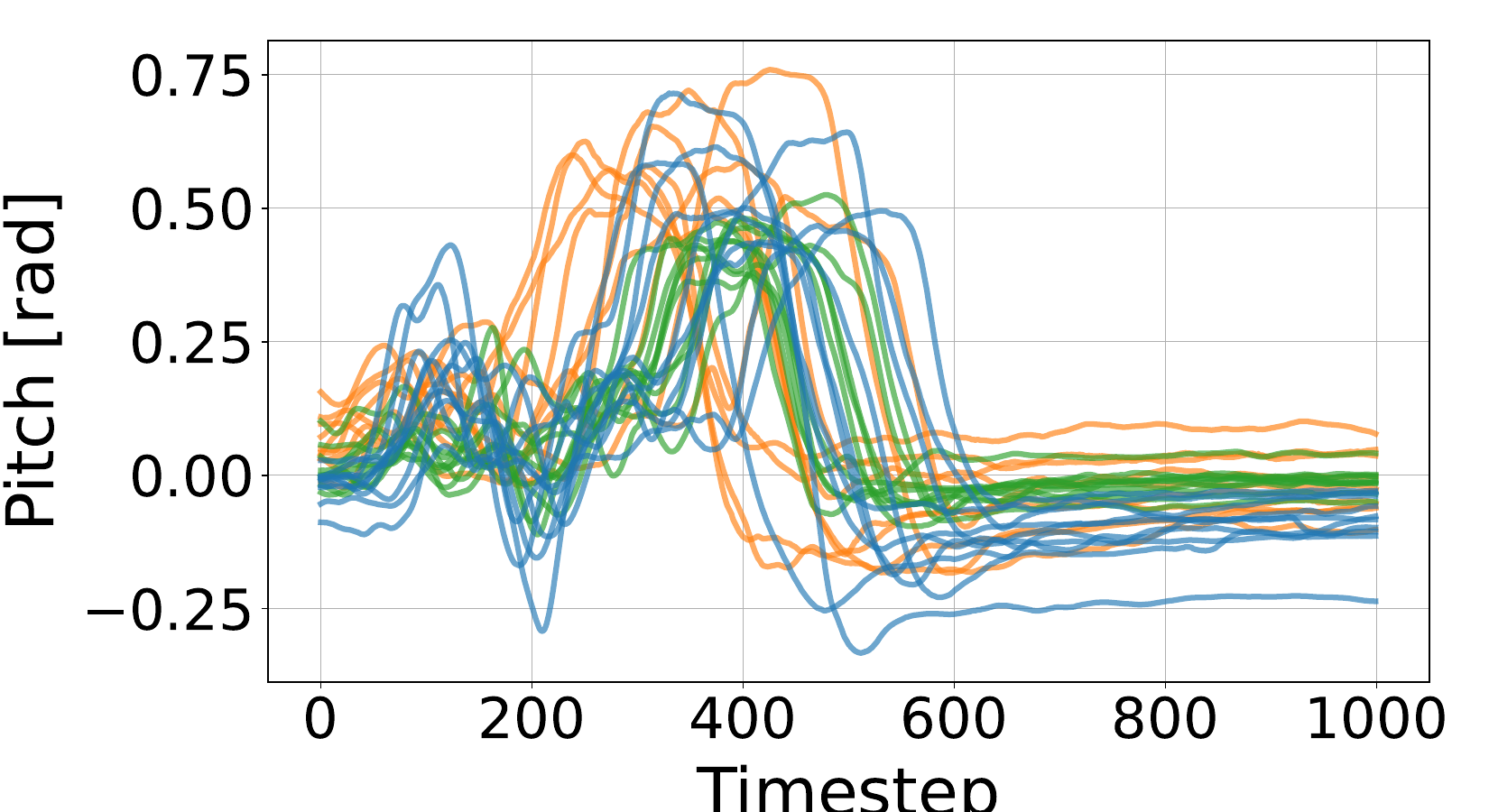}
        \subcaption{Pitch trajectories}
        \label{fig:human_demo_pitch}
    \end{subfigure}
    \caption{
    Human-collected object-reaching-and-picking demonstrations used in the quadruped robot experiment.
    The demonstrations include variations in planar position and pitch motion.
    }
    \label{fig:human_demo}
\end{figure}

%%%%%%%%%%%%%%%%%%%%%%%%%%%%%%%%%%%%%%%%%%%%%%%%%%%%%%%%%%%%%%

To address RQ3, we evaluate EF-GAIfO with human-collected demonstrations using a Unitree Go2 quadruped robot in the Isaac Sim physics simulator.
% To evaluate EF-GAIfO with human-collected demonstrations, we conduct experiments using a Unitree Go2 quadruped robot in the Isaac Sim physics simulator.
The task is an object-reaching-and-picking task, in which the robot moves from the initial position toward a target object and picks it up using the gripper (Fig. \ref{fig:go2_task_setting}).
The robot starts from the initial position \((x,y)=(0,0)\) m.
Human demonstrations are collected for three target object positions: \((2.25,0.0)\) m, \((2.0,-0.25)\) m, and \((2.0,0.25)\) m.
An episode terminates when the maximum episode length is reached or when the robot base contacts the ground. A trial is considered successful if the robot grasps and lifts the target object before termination.

The human-collected demonstrations intentionally include challenging motions, such as fast reaching motions and large pitch-angle motions, which may be difficult for the robot to reproduce. 
Figure~\ref{fig:robot:experient:demo} shows the human demonstration setup used in the robot experiment.
The object-reaching-and-picking motion is recorded using a robot-free demonstration interface.
The dataset contains 30 trajectories, with 10 trajectories collected for each of the three target positions.
Each trajectory lasts approximately \(10\,\mathrm{s}\) and is recorded at \(100\,\mathrm{Hz}\).
Figure~\ref{fig:human_demo} visualizes the collected demonstrations, including planar position and pitch trajectories.

We use a hierarchical policy structure consisting of a pretrained low-level locomotion policy and a learned high-level policy.
The low-level policy controls the whole-body locomotion of the robot, while the high-level policy generates commands for navigation, body pitch, and gripper motion.
The high-level action consists of forward velocity, lateral velocity, yaw rate, pitch command, and gripper open/close command.
The observation of the high-level policy consists of the target position relative to the robot frame \((x,y)\), robot base linear and angular velocities, robot orientation, gripper pitch angle, and gripper open/close state.

% The high-level policy is trained for 3,000 iterations.
% The reconstruction model is implemented as a VAE with a latent dimension of eight.

The high-level policy is trained for 3000 iterations, with 12 interaction steps collected from each environment per iteration.
The actor, critic, and discriminator networks are implemented as multilayer perceptrons with two hidden layers of 512 and 256 units, respectively.
The reconstruction model is implemented as a variational autoencoder.
Its encoder and decoder use two hidden layers of 128 and 64 units, and the latent dimension is set to eight. All hidden layers use ReLU activation functions, and all networks are optimized using Adam.
We compare EF-GAIfO with GAIfO and WGAIfO under the same policy-learning settings.

%%%%%%%%%%%%%%%%%%%%%%%%%%%%%%%%%%%%%%%%%%%%%%%%%%%%%%%%
\subsubsection{Results}
% both and WGAIfO  

\begin{table}[t]
    \centering
    \caption{
        Object-reaching-and-picking task success rates in simulation for three fixed target positions and a random target setting.
        % Each setting is evaluated over 100 trials.
        Each value is the mean success rate over five policies trained with different random seeds, with each policy evaluated over 100 trials.
    }
    \label{tab:go2_results}
    \begin{tabular}{cccc}
        \toprule
                        & EF-GAIfO                 & WGAIfO        & GAIfO \\
        \midrule
            Target 1 & $\mathbf{83.6 \pm 9.2\%}$  & $34.8 \pm 47.7\%$ & $37.4 \pm 14.2\%$ \\
            Target 2        & $\mathbf{88.0 \pm 5.7\%}$  & $35.2 \pm 47.8\%$ & $36.8 \pm 8.1\%$ \\
            Target 3        & $\mathbf{84.8 \pm 10.0\%}$ & $39.0 \pm 48.8\%$ & $42.0 \pm 17.6\%$ \\
            Random target   & $\mathbf{85.8 \pm 7.4\%}$  & $26.4 \pm 44.1\%$ & $35.4 \pm 18.1\%$ \\
        \bottomrule
    \end{tabular}
\end{table}

%%%%%%%%%%%%%%%%%%%%%%%%%%%%%%%%%%%%%%%%

\begin{figure}[t]
    \centering

    % (a) EF-GAIfO: top middle
    \begin{minipage}[b]{0.48\linewidth}
        \centering
        \includegraphics[width=\linewidth]{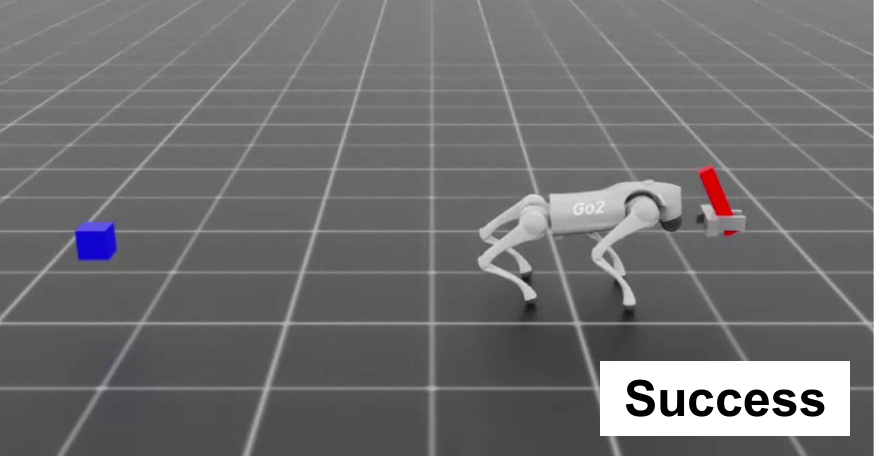}
        \subcaption{EF-GAIfO}
        \label{fig:go2:result:efgaifo}
    \end{minipage}

    \vspace{2mm}

    % (b) WGAIfO: bottom left
    \begin{minipage}[b]{0.48\linewidth}
        \centering
        \includegraphics[width=\linewidth]{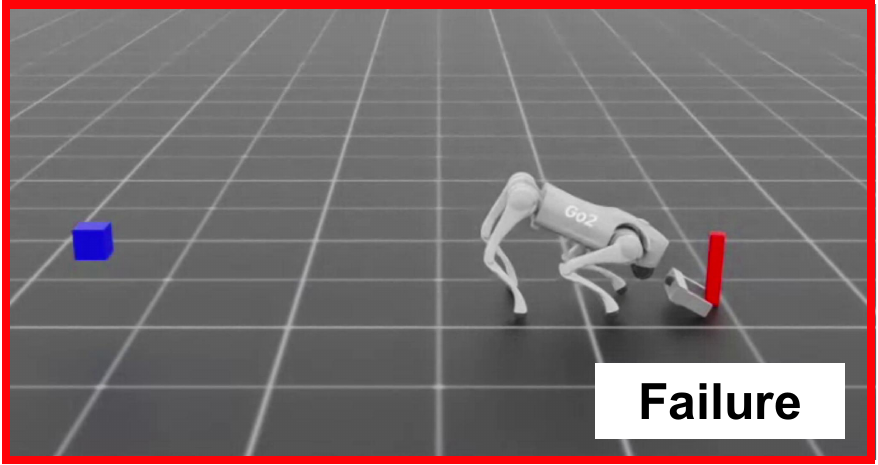}
        \subcaption{WGAIfO}
        \label{fig:go2:result:wgaifo}
    \end{minipage}
    \hfill
    % (c) GAIfO: bottom right
    \begin{minipage}[b]{0.48\linewidth}
        \centering
        \includegraphics[width=\linewidth]{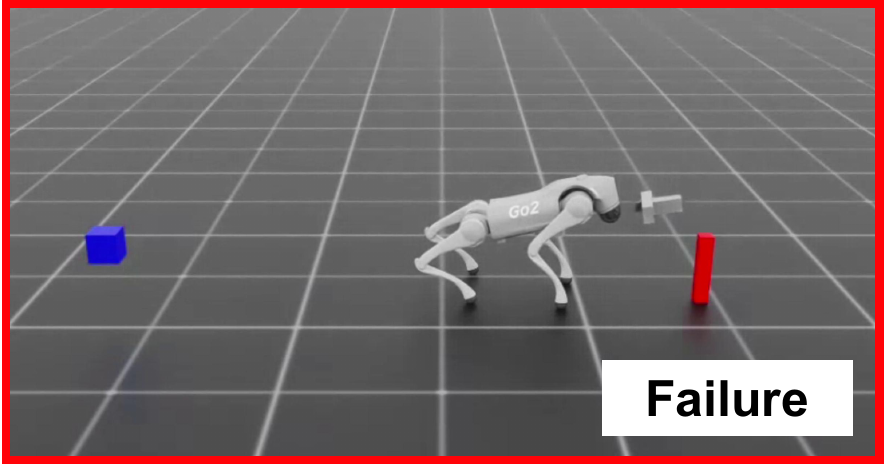}
        \subcaption{GAIfO}
        \label{fig:go2:result:gaifo}
    \end{minipage}

    \caption{
         Execution of the learned policies in the object-reaching-and-picking task using the Go2 robot in simulation.
         (a) EF-GAIfO successfully completes the task.
         (b) WGAIfO fails due to excessive pitch motion.
         (c) GAIfO fails due to misalignment between the gripper and the object.
    }
    \label{fig:go2:result}
\end{figure}
%%%%%%%%%%%%%%%%%%%%%%%%%%%%%%%%%%%%%%%%%

\begin{figure}[t]
    \centering

    % Feasibility weight colorbar
    \includegraphics[width=0.7\linewidth]
    {figure/feasibility_weight_colorbar.pdf}
    \vspace{1mm}
    \begin{minipage}[b]{0.48\linewidth}
        \centering
        \includegraphics[width=1\linewidth]{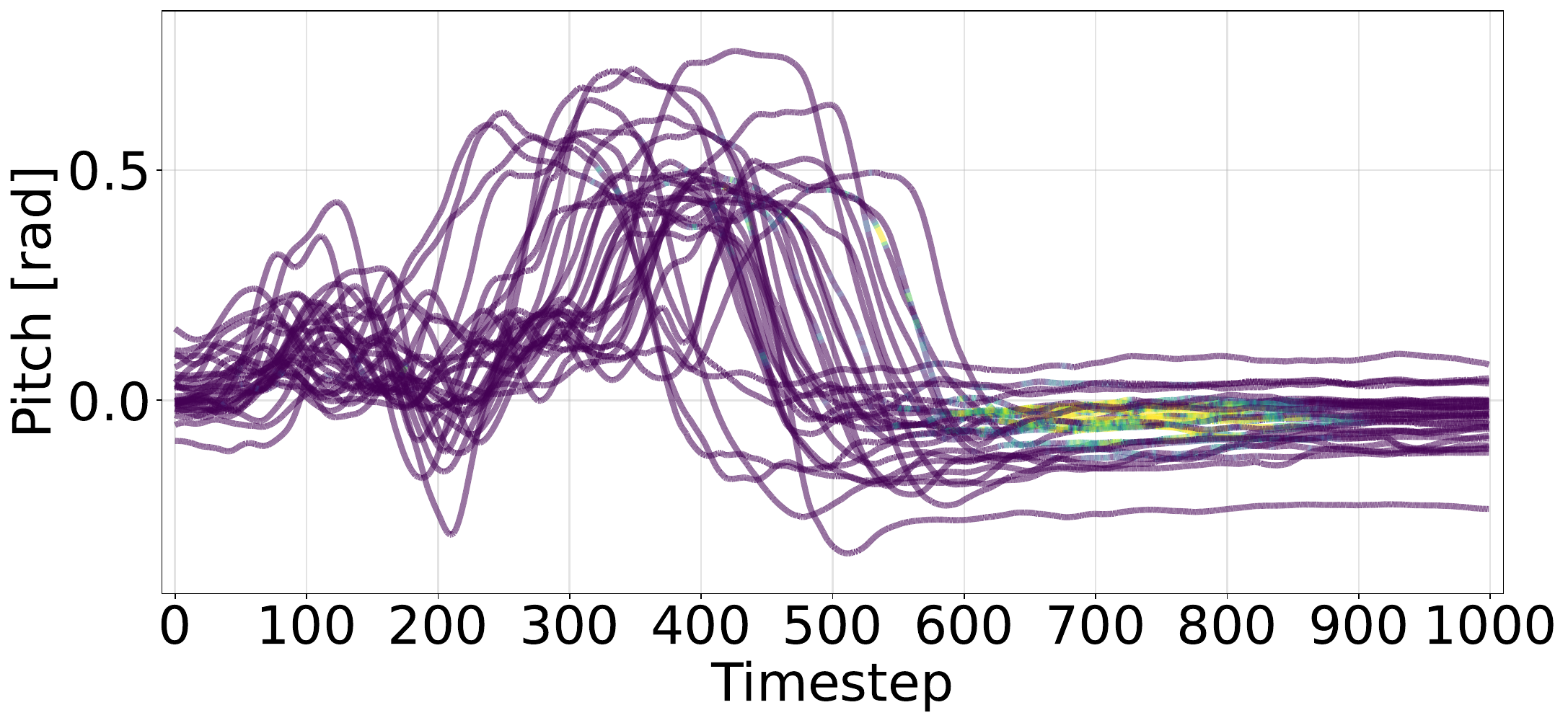}
        \subcaption{100 iter.}
        \label{fig:go2:feasibility_100}
    \end{minipage}
    \begin{minipage}[b]{0.48\linewidth}
        \centering
        \includegraphics[width=1\linewidth]{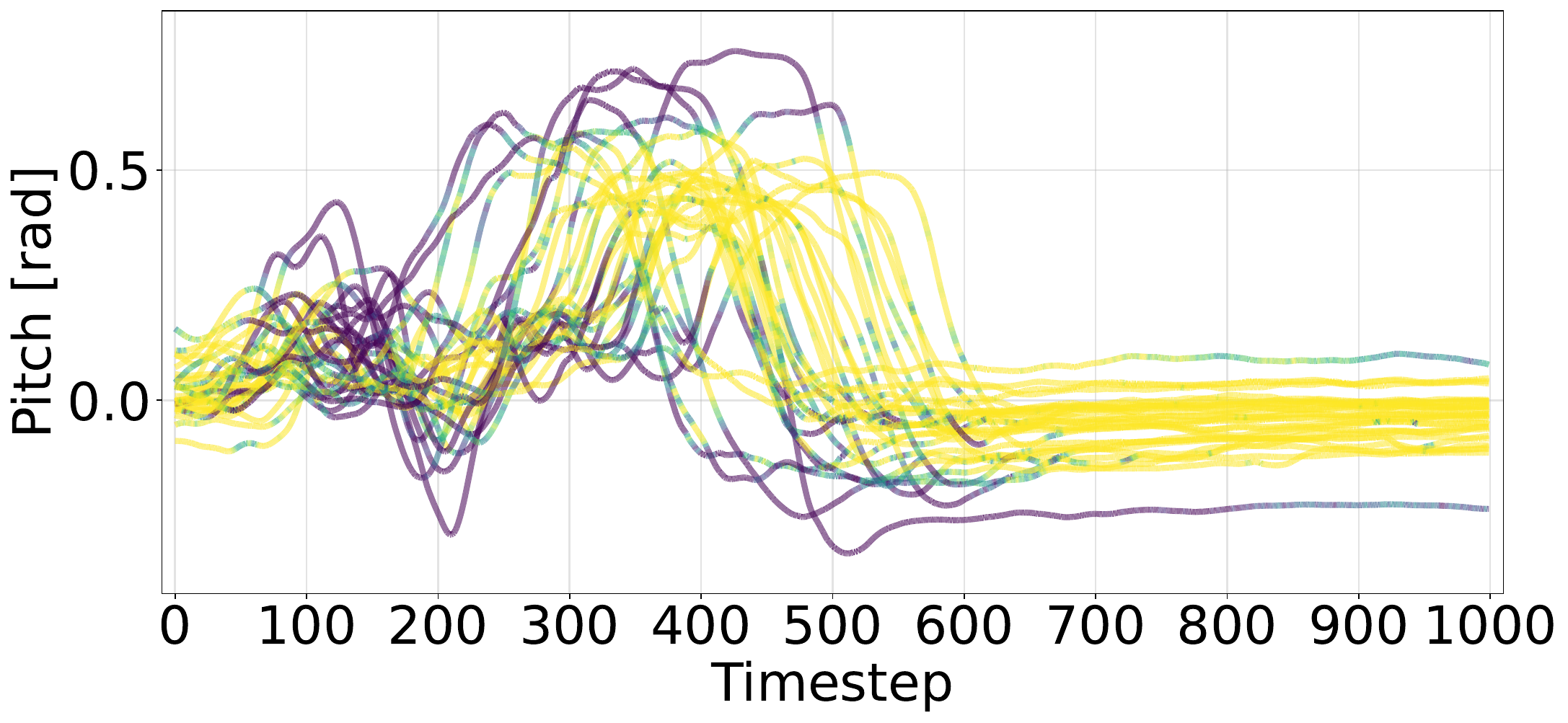}
        \subcaption{3000 iter.}
        \label{fig:go2:feasibility_3000}
    \end{minipage} 
    % \caption{Feasibility weights estimated by EF-GAIfO for pitch motions in the object-picking simulation environment.}
    \caption{
        Feasibility weights estimated by EF-GAIfO for demonstrated pitch motions in the quadruped robot experiment at (a) 100 and (b) 3000 training iterations.
    }
    \label{fig:go2:result:feasibility}
\end{figure}

We evaluate object-picking performance using human-collected demonstrations in simulation.
Table~\ref{tab:go2_results} reports success rates for three fixed target positions and a random target setting, where the target is sampled from \(x \in [2.0, 2.5]\) m and \(y \in [-0.25, 0.25]\) m.
EF-GAIfO achieves higher success rates across all settings, outperforming GAIfO and WGAIfO.
These results show that EF-GAIfO improves task performance and generalizes to nearby target positions.

% Figure~\ref{fig:go2:result} shows the simulated Go2 robot controlled by the policy learned by each method.
% The policy learned by EF-GAIfO more reliably approaches the object and completes the pick-up behavior.
% In contrast, the WGAIfO policy generates large pitch commands, causing strong contact between the gripper and the ground.
% The GAIfO policy shows failure cases where the gripper position is misaligned with the object.
% For WGAIfO, this gripper-ground contact prevents the gripper from closing and causes object-picking failure. GAIfO instead often fails because the gripper is misaligned with the object. 

Figure~\ref{fig:go2:result} shows the simulated Go2 robot controlled by the policy learned by each method.
The policy learned by EF-GAIfO more reliably approaches the object and completes the pick-up behavior.
In contrast, the WGAIfO policy generates large pitch commands, causing strong contact between the gripper and the ground.
This gripper-ground contact prevents the gripper from closing and causes object-picking failure. 
The GAIfO policy shows failure cases where the gripper position is misaligned with the object.

Figure~\ref{fig:go2:result:feasibility} shows the estimated feasibility weights for the demonstrated pitch motions.
As training progresses, large-pitch transitions receive lower weights, indicating that EF-GAIfO suppresses motions that are difficult for the robot to reproduce.

These simulation results address the learning component of RQ3. EF-GAIfO successfully learns an object-picking policy from human-collected state-only demonstrations and consistently achieves higher success rates than GAIfO and WGAIfO across both fixed and random target settings. The following experiment examines whether the learned policy can also be successfully executed on a real robot.

%%%%%%%%%%%%%%%%%%%%%%%%%%%%%%%%%%%%%%%%%%%%%%%%%%%%%%%%%%%%%%%%%%%%%%%%%%%%%%%

\subsubsection{Real-World Experiment}
\label{subsubsec:real_world_experiment}

% After the simulation experiments, we deploy the learned policy trained in simulation on the real Unitree Go2 robot to evaluate its real-world executability.
% We use the same object-picking task and evaluate three fixed target positions and a random target setting, with 10 trials per setting.
% Table~\ref{tab:real_go2_results} reports the real-world success rates, and Fig.~\ref{fig:real_robot_motion} shows the robot executing the learned high-level policies.
% EF-GAIfO policy successfully approaches, grasps, and lifts the target object while maintaining stable locomotion, demonstrating that the policy learned in simulation can be transferred to the real robot.
% 
% Together with the simulation results, the real-world evaluation answers RQ3. EF-GAIfO learns effectively from human-collected state-only demonstrations, and the resulting policy can be successfully executed on the real robot, achieving success rates of 70–80\% across the fixed and random target settings. These results demonstrate the real-world executability of the behavior learned using EF-GAIfO.

We deploy the policy trained in simulation on the real Unitree Go2 robot to evaluate its real-world executability.
We use the same object-picking task and evaluate three fixed target positions and a random target setting, with ten trials per setting.

% Table~\ref{tab:real_go2_results} reports the success rates, and Fig.~\ref{fig:real_robot_motion} shows the robot executing the learned high-level policies.
% EF-GAIfO policy successfully approaches, grasps, and lifts the target object while maintaining stable locomotion, demonstrating that the policy learned in simulation can be transferred to the real robot.

\begin{figure*}[t]
    \centering
    \begin{subfigure}[t]{\linewidth}
        \centering
        \includegraphics[width=\linewidth]{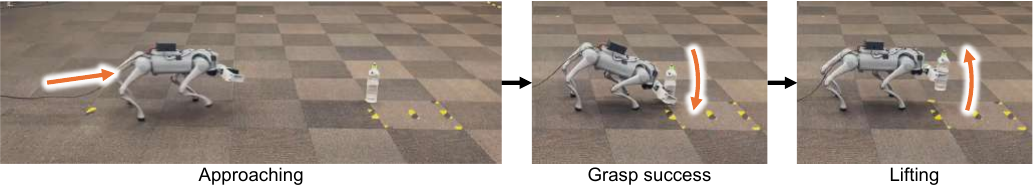}
        \subcaption{EF-GAIfO}
        \label{fig:real_robot_motion:ef-gaifo}
    \end{subfigure}
    \vspace{1mm}
    \begin{subfigure}[t]{0.48\linewidth}
        \centering
        \includegraphics[width=\linewidth]{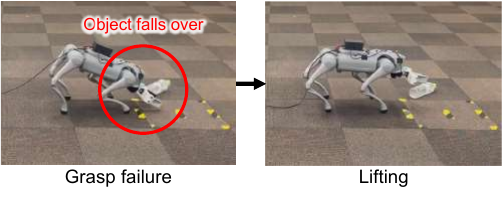}
        \subcaption{WGAIfO}
        \label{fig:real_robot_motion:wgaifo}
    \end{subfigure}
    \hfill
    \begin{subfigure}[t]{0.48\linewidth}
        \centering
        \includegraphics[width=\linewidth]{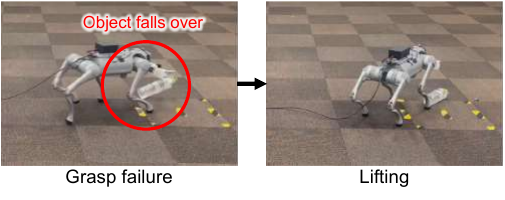}
        \subcaption{GAIfO}
        \label{fig:real_robot_motion:gaifo}
    \end{subfigure}
    \caption{
        Representative real-world execution examples of the learned policies in the object-reaching-and-picking task using the Go2 robot.
        The snapshots illustrate typical behaviors observed for each method:
        (a) EF-GAIfO successfully approaches, grasps, and lifts the object;
        (b) WGAIfO exhibits a grasping failure;
        and (c) GAIfO exhibits a grasping failure caused by gripper–object misalignment.
        % Real-world execution of the learned EF-GAIfO policy on the Unitree Go2 robot in the object-picking task.
        % Real-world execution of the learned policies in the object-reaching-and-picking task using the Go2 robot.
        % (a) EF-GAIfO successfully completes the task.
        % (b) WGAIfO fails to grasp the object.
        % (c) GAIfO fails to grasp the object.
    }
    \label{fig:real_robot_motion}
\end{figure*}

Table~\ref{tab:real_go2_results} reports the success rates, and Fig.~\ref{fig:real_robot_motion} shows representive execution examples of the learned policies on the actual Go2 robot.
EF-GAIfO achieves consistently higher success rates across all target settings and successfully approaches, grasps, and lifts the target object while maintaining stable locomotion.
In contrast, WGAIfO and GAIfO frequently fail during the grasping phase, resulting in substantially lower task success rates.
These results indicate that the policy learned with EF-GAIfO in simulation can be transferred to the actual robot and can reliably execute the object-reaching-and-picking behavior under both fixed and randomly sampled target positions.

Together with the simulation results, the real-world evaluation answers RQ3. EF-GAIfO learns effectively from human-collected state-only demonstrations, and the resulting policy can be successfully executed on the real robot, achieving success rates of 70–80\% across the fixed and random target settings. These results demonstrate the real-world executability of the behavior learned using EF-GAIfO.

% \begin{table}[t]
%     \centering
%     \caption{Real-world object-picking success rates of learned policies. Each setting is evaluated over 10 trials.}
%     \label{tab:real_go2_results}
%     \renewcommand{\arraystretch}{1.15}
%     \begin{tabular}{ccccc}
%         \toprule
%                   & Target 1 & Target 2 & Target 3 & Random target   \\
%         \midrule
%         EF-GAIfO  & \textbf{70} & \textbf{70} & \textbf{80} & \textbf{80} \\
%         WGAIfO   & 10       & 10      & 10      & 30 \\
%         GAIfO     & 10       & 10      & 20      & 20 \\
%         \bottomrule
%     \end{tabular}
% \end{table}

\begin{table}[t]
    \centering
    \caption{
        Real-world object-reaching-and-picking task success rates for three fixed target positions and a random target setting. Each setting is evaluated over 10 trials.
    }
    \label{tab:real_go2_results}
    \begin{tabular}{cccc}
        \toprule
                        & EF-GAIfO      & WGAIfO & GAIfO \\
        \midrule
        Target 1        & \textbf{70\%}   & 10\%     & 10\% \\
        Target 2        & \textbf{70\%}   & 10\%     & 10\% \\
        Target 3        & \textbf{80\%}   & 10\%     & 20\% \\
        Random target   & \textbf{80\%}   & 30\%     & 20\% \\
        \bottomrule
    \end{tabular}
\end{table}

%%%%%%%%%%%%%%%%%%%%%%%%%%%%%%%%%%%%%%%%%%%%%%%%%%%%%%%%%%%%%%%%%%%%%%%%%%%%%%%

% \section{Discussion}

%%%%%%%%%%%%%%%%%%%%%%%%%%%%%%%%%%%%%%%%%%%%%%%%%%%%%%%%%%%%%%%%%%%%%%%%%%%%%%%
\section{Conclusions and Future Work}

We proposed EF-GAIfO, an experience-based feasibility-aware framework for imitation from state-only demonstrations that may contain robot-infeasible transitions. The experiments showed that the feasibility estimates became more discriminative as robot experience accumulated and that EF-GAIfO achieved higher task success rates and closer trajectory matching across different forms of feasibility mismatch. With human-collected demonstrations, EF-GAIfO also achieved high success rates in simulation and successful execution on a real quadruped robot. These results demonstrate the effectiveness and practical applicability of experience-based feasibility estimation for demonstrations with unknown feasibility. 

% Future work includes extending feasibility estimation to environmental constraints and expanding EF-GAIfO to other observation modalities. Regarding environmental constraints, this study mainly considers infeasible state transitions caused by robot's embodiment. However, in more complex environments, feasibility can also depend on obstacles, object placement, and contact conditions. Developing a framework that estimates feasibility by considering both robot's embodiment constraints and environmental conditions is an important direction for future work. In this study, EF-GAIfO is applied to state-transition observations obtained from a robot-free demonstration interface. However, robot-free demonstration interfaces can provide various types of observations, such as images, videos, depth information, and sensor measurements. Extending experience-based feasibility estimation to these observation modalities would make EF-GAIfO applicable to a wider range of human demonstration settings.

The EF-GAIfO estimates feasibility from robot-experienced state transitions and therefore mainly captures embodiment-related motion feasibility.
However, in real-world tasks, whether a transition is feasible can also depend on the surrounding environment, such as terrain, obstacles, object placement, and contact conditions.
Addressing such environment-dependent feasibility requires extending the current estimator to condition feasibility not only on robot motion but also on environmental context.
Future work will therefore investigate feasibility estimation that jointly represents robot state transitions and environmental observations, including visual, depth, and other sensory information.

%%%%%%%%%%%%%%%%%%%%%%%%%%%%%%%%%%%%%%%%%%%%%%%%%%%%%%%%%%%%%%%%%%%%%%%%%%%%%%%%

%%%%%%%%%%%%%%%%%%%%%%%%%%%%%%%%%%%%%%%%%%%%%%%%%%%%%%%%%%%%%%%%%%%%%%%%%%%%%%%%

%%%%%%%%%%%%%%%%%%%%%%%%%%%%%%%%%%%%%%%%%%%%%%%%%%%%%%%%%%%%%%%%%%%%%%%%%%%%%%%%
\bibliographystyle{IEEEtran}
\small
\bibliography{reference}
\normalsize
%%%%%%%%%%%%%%%%%%%%%%%%%%%%%%%%%%%%%%%%%%%%%%%%%%%%%%%%%%%%%%%%%%%%%%%%%%%%%%%%

\end{document}